\documentclass[11pt]{article}

\usepackage[final]{acl}

\usepackage{times}
\usepackage{latexsym}
\usepackage{multirow}
\usepackage{amssymb}
\usepackage{enumitem}

\newcommand{\inlinecode}[1]{\texttt{#1}}

\usepackage[T1]{fontenc}

\usepackage[utf8]{inputenc}

\usepackage{microtype}

\usepackage{inconsolata}

\usepackage{graphicx}

\usepackage{microtype}
\usepackage{hyperref}
\usepackage{url}
\usepackage{booktabs}
\usepackage{amsmath}
\usepackage{graphicx}
\usepackage{makecell}
\usepackage[table]{xcolor}
\usepackage{colortbl}
\usepackage{tcolorbox}
\usepackage{xspace}
\usepackage{nicematrix}

\newtcolorbox{planbox}[1]{
    colback=gray!5,
    colframe=gray!75,
    title=#1,
    fonttitle=\bfseries
}

\definecolor{genLLM}{RGB}{230,230,230} 
\definecolor{specLLM}{RGB}{252,228,214}  
\definecolor{ourLLM}{RGB}{230,242,230}  
\definecolor{spaLLM}{RGB}{255,249,196}  
\definecolor{genVLM}{RGB}{255,255,255}   
\definecolor{ourVLM}{RGB}{207,234,242} 

\newcommand{\benchname}{Vbvr-VQA benchmark\xspace}
\newcommand{\modelname}{ChronoVision\xspace}

\definecolor{line-blue}{RGB}{243, 248, 252}

\usepackage{lineno}

\definecolor{darkblue}{rgb}{0, 0, 0.5}
\hypersetup{colorlinks=true, citecolor=darkblue, linkcolor=darkblue, urlcolor=darkblue}

\title{ChronoVision: Temporal Reasoning via Latent State Reconstruction}

\author{
\textbf{Yifan Shen\textsuperscript{1,2}} \quad
\textbf{Jian Xu\textsuperscript{1}} \quad
\textbf{Boyi Li\textsuperscript{1}} \quad
\textbf{Yuner Zhang\textsuperscript{3}} \quad
\textbf{Tianjiao Yu\textsuperscript{1}} \\[3pt]
\textbf{Bingxuan Li\textsuperscript{1}} \quad
\textbf{Houze Yang\textsuperscript{1}} \quad
\textbf{Rushi Wang\textsuperscript{1}} \quad
\textbf{Xu Cao\textsuperscript{1,2}} \\[3pt]
\textsuperscript{1}University of Illinois Urbana-Champaign \quad
\textsuperscript{2}PediaMed AI \quad
\textsuperscript{3}University of Pennsylvania \\
\texttt{yifan26@illinois.edu}
}

\begin{document}
\maketitle
\begin{abstract}
Multimodal large language models excel at passive perception but struggle with complex visual cognitive tasks requiring multi-step temporal reasoning. This degradation largely stems from the inherent ambiguity of language-based reasoning, which often fails to accurately articulate continuous visual transformations. To address this, we propose ChronoVision, a multimodal framework designed to align visual logic with latent imagery. During supervised fine-tuning, a Reconstructive Visual Head predicts the latent representation of the final transformed state, while an ROI Attention Locating module focuses the model on key visual evidence via semantic span queries. In post-training, we apply reinforcement learning with an implicit process grounding mechanism, guided by a composite reward function that evaluates outcome correctness, latent process alignment, and unsupervised visual focus. Furthermore, we introduce Vbvr-VQA, a novel dataset that evaluates temporal tracking by reformulating video reasoning into a strict image-ordering task. Experiments demonstrate that ChronoVision achieves state-of-the-art performance on Vbvr-VQA with 74.8\% in-domain and 71.6\% out-of-domain accuracy, alongside a strong 55.0\% accuracy on IntPhys2, a highly challenging cross-domain benchmark. \url{https://pediamedai.com/Cognition-MLLM/ChronoVision/}
\end{abstract}

\section{Introduction}
Multimodal Large Language Models have achieved remarkable success in open vocabulary perception tasks such as image captioning and object detection \citep{liu2023visual}. However, their capabilities remain highly limited when faced with complex cognitive tasks that require spatial intelligence~\citep{zhang2026theory} and visual reasoning \citep{li2025coreknowledgedeficitsmultimodal, schulze2025visual}. In contrast, human cognition intuitively bridges this gap through mental simulation. When observing physical events, such as an object rotating in space, humans can effortlessly project spatial trajectories in brain and predict continuous visual transformations. This cognitive ability to internally simulate and predict the evolution of visual states is known as visual imagery. Current models struggle with tasks that demand visual memory and multiple-step temporal reasoning, in which they rely on recognizing static visual content rather than understanding continuous space with real physical properties.

Current mainstream approaches attempt to enhance reasoning capabilities by extending Chain of Thought reasoning within the textual space \citep{wei2023chainofthoughtpromptingelicitsreasoning, zhang2024multimodalchainofthoughtreasoninglanguage}. This creates a fundamental bottleneck. Natural language cannot precisely articulate continuous visual transformations. Attempting to verbally describe a complex three-dimensional rotation or a continuous physical motion trajectory inevitably leads to the loss of critical spatial information \citep{schulze2025visual}. Therefore, purely text-based reasoning is inadequate for solving cognitive tasks that rely heavily on visual imagery \citep{qin2025chain}.

Deeper analysis also reveals significant flaws in existing evaluation methods \citep{cao2024visual, khezresmaeilzadeh2026vriq}. Traditional Visual Question Answering tasks typically adopt a multiple-choice format. This design introduces a critical issue. The textual descriptions in the question stem or options can be ambiguous, allowing models to exploit linguistic pattern-matching rather than performing authentic visual reasoning. A model might guess the correct option based on text semantics without referencing the visual observation of the transformed result. Furthermore, selecting a single final answer fails to evaluate temporal sequence reasoning capabilities. Genuine visual reasoning requires tracking the evolutionary trajectory from an initial state, through intermediate states, to a final outcome.

To address these limitations, we propose a rigorous new evaluation paradigm and a novel multimodal reasoning framework. First, we present Vbvr-VQA, a benchmark that reformulates video reasoning as a strict image-ordering task \citep{wang2026bigvideoreasoningsuite}. Instead of allowing free-form language outputs, this task requires a model to arrange shuffled candidate frames into their correct chronological order, reducing opportunities for linguistic shortcuts and forcing a genuine understanding of temporal progression. To tackle this challenging setting, we then introduce ChronoVision. Its central idea is to encourage the model to internally reconstruct the visual outcome of an event within its latent space, mirroring the human tendency to mentally imagine how a scene should unfold \citep{lee2025perspectiveawarereasoningvisionlanguagemodels}.

The training pipeline of ChronoVision consists of two main stages. During the supervised fine-tuning stage, we introduce the Reconstructive Visual Head. This module trains the model to use scattered information from the candidate frames to predict the latent representation of the final transformed state. Simultaneously, we incorporate a Region of Interest Attention Locating module. Guided by semantic locating cues, this module directs the model to precisely focus on the fine-grained dynamic key regions required for visual reasoning. Given long-horizon reasoning often leads to compounding errors \citep{motwani2025h1bootstrappingllmsreason, li2026limitedreasoningspacecage,li2025exploring,li2025stitchfusion,li2025maris,li2025exploring}, we implement a reinforcement learning stage based on implicit process grounding. We design a Composite Reward Function that simultaneously evaluates outcome correctness, latent visual alignment, and attention focus. ChronoVision achieves a 74.8 percent in domain accuracy and a 71.6 percent out-of-domain accuracy on Vbvr-VQA, outperforming all compared open-source and proprietary models. Furthermore, ChronoVision achieves state-of-the-art performance on the challenging IntPhys2 benchmark \citep{bordes2025intphys}, with an overall accuracy of 55.0\%. Overall, our contributions are summarized as follows:
\begin{itemize}
\item First, we introduce Vbvr-VQA. We redesign the video reasoning task as an image ordering task. Given an initial frame, a text prompt, and six shuffled frames, the model is required to accurately order the sequence. This forces the model to genuinely comprehend temporal evolution rather than relying on linguistic shortcuts.
\item Second, we propose ChronoVision, a comprehensive framework for visual temporal reasoning. We introduce the Reconstructive Visual Head to predict the visual representation of the final transformed state within the internal latent space. We also introduce the Region of Interest Attention Locating module to guide the focus toward fine-grained dynamic key regions.
\item Third, we incorporate a reinforcement learning stage based on implicit process grounding. We design a Composite Reward Function that comprehensively evaluates outcome correctness, latent visual alignment, and attention focus. This approach optimizes the internal reasoning trajectory and significantly improves robustness in long-horizon tasks.
\item Fourth, extensive experiments validate the superiority of our approach. ChronoVision achieves state-of-the-art performance on Vbvr-VQA and attains the best overall accuracy on the challenging cross-domain IntPhys2 benchmark \citep{bordes2025intphys}.
\end{itemize}
\section{Dataset and Benchmark}

\begin{figure*}[t]
    \centering
    \includegraphics[width=\linewidth]{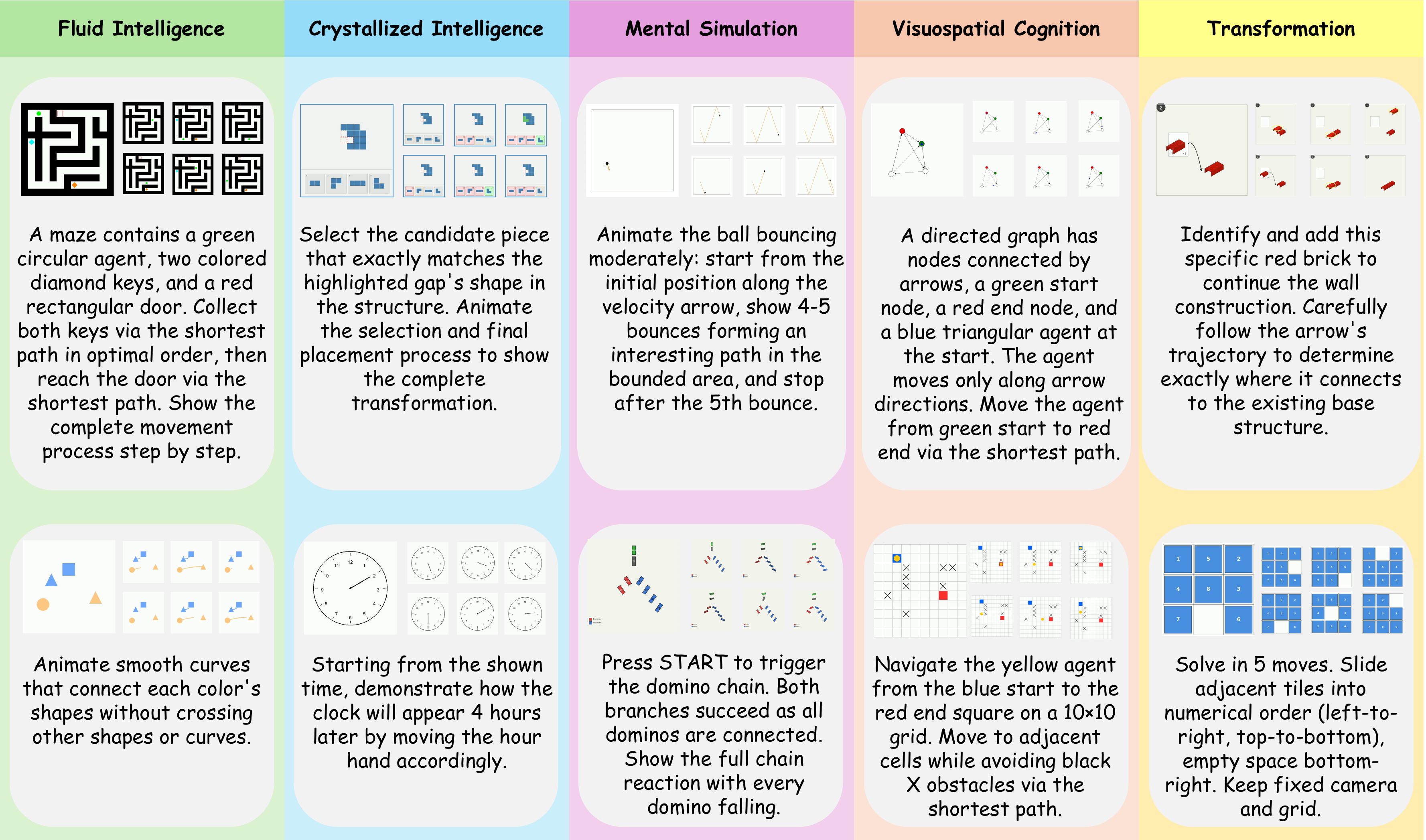} 
    \caption{Examples of the Vbvr-VQA dataset}
    \label{fig:MainBaby_fig}
\end{figure*}

Current Visual Question Answering benchmarks primarily assess static image and text alignment alongside fundamental visual recognition. The community currently lacks datasets designed to evaluate abstract visual reasoning, including multistep transformation inference, causal temporal dynamics, and complex scene understanding. To address this limitation, we introduce Vbvr-VQA, built upon the open source Very Big Video Reasoning dataset \citep{wang2026bigvideoreasoningsuite}. These samples span diverse cognitive task categories and are partitioned into in domain and zero shot generalization splits. To explicitly measure the temporal logic and state tracking capabilities of the models, we reformulate the traditional multiple choice format into a strict cognitive visual sequence task.

This reformulation into an image-ordering task is a deliberate design choice rather than a limitation. Traditional visual question answering benchmarks frequently rely on multiple-choice formats or free-form text generation. These formats contain a critical flaw: models can exploit linguistic pattern-matching within the question stems or options to bypass actual visual comprehension. By enforcing a text-free, strict ordering formulation, we eliminate linguistic shortcuts and force the model to execute pure temporal and physical causal reasoning. Furthermore, while the output format is unified, the underlying physical scenarios are extremely diverse. The benchmark incorporates 100 distinct task generators covering fluid dynamics, kinematic collisions, and continuous spatial rotations. Testing models through a unified, strictly constrained interface across highly heterogeneous physical domains represents a standard paradigm for rigorous cognitive evaluation.

Under the standard evaluation pipeline, the model receives an initial video frame and a text prompt that serve as the visual premise and reasoning instruction. The model is then provided with six shuffled candidate images representing different evolutionary states of the visual transformation. The core objective is to predict the exact chronological permutation of these unordered frames. To construct this task, we uniformly divide the ground truth video along the temporal axis into six intervals of equal length. We extract the final frame of each segment to form a temporal sequence that depicts the progressive transformation. We subsequently apply a random permutation to these frames to generate the unordered candidate set.

To support fine grained visual reasoning and mitigate the common grounding perception mismatch problem in Vision Language Models, we augment Vbvr-VQA with dense annotations. Based on the ground truth sequence, the annotation pipeline synthesizes intermediate reasoning traces to explicitly identify the dynamic transformation regions. This process generates semantic visual cues formatted as \texttt{<LOCATE>...</LOCATE>}. Simultaneously, the pipeline provides precise bounding box coordinates for the regions undergoing state changes. These coordinates serve as dense spatial supervision for the downstream attention modules. Figure \ref{fig:MainBaby_fig} presents a set of representative examples from the Vbvr-VQA benchmark to illustrate the diversity of the visual reasoning tasks. By formulating the evaluation as a sequence ordering task rather than a single choice selection, Vbvr-VQA strictly requires the model to understand physical trajectories and causal relationships. This design effectively prevents the models from exploiting shortcuts based on superficial single frame matching. We provide additional information in Appendix \ref{appendix:dataset_vis}.
\section{Methodology}

\subsection{Preliminaries and Problem Formulation}

\begin{figure*}[t]
    \centering
    \includegraphics[width=\textwidth]{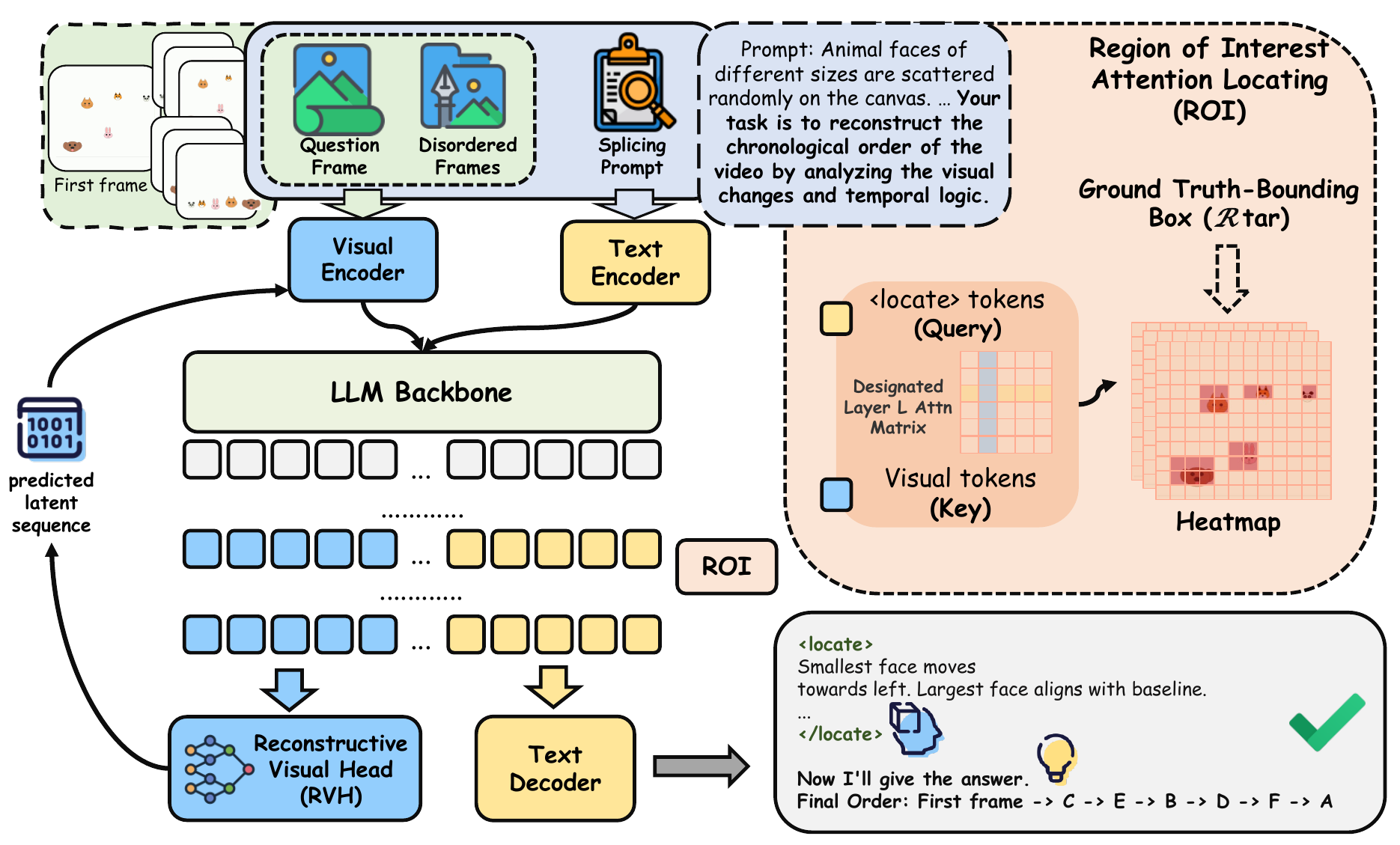}
    \caption{Overall pipeline of ChronoVision. Given a query frame, shuffled candidate frames, and a task prompt, the model encodes visual and textual inputs with a unified backbone. On top of this backbone, the Reconstructive Visual Head predicts the latent representation of the final transformed state, while the ROI Attention Locating module guides attention toward dynamic evidence regions for temporal reasoning.}
    \label{fig:method_pipeline}
\end{figure*}

We aim to train a multimodal large language model, parameterized by $\theta$, to model the conditional distribution $\pi_\theta(y \mid x)$ for a multimodal input $x = (\mathcal{V}, \mathcal{Q})$. Here, $y$ is the predicted chronological permutation. The visual input $\mathcal{V}$ includes an initial query image $v_{\text{query}}$ and a candidate set $\mathcal{O} = \{o_1, \dots, o_6\}$ sampled from distinct temporal points. The textual prompt $\mathcal{Q}$ specifies the reasoning task. A visual encoder $\text{Encoder}_{\text{vis}}(\cdot)$ extracts dense features from $\mathcal{V}$, which are projected into visual tokens. The language model backbone $\text{Encoder}_{\text{txt}}(\cdot)$ processes the concatenated visual and text tokens to generate $y$ autoregressively.

Figure~\ref{fig:method_pipeline} provides an overview of ChronoVision. We next describe its three main components: the Reconstructive Visual Head, the ROI Attention Locating module, and the reinforcement learning stage with implicit process grounding.

\subsection{Supervised Fine-Tuning with Reconstructive Visual Head}
\label{method_subsec:RVH}

To acquire advanced visual cognitive capabilities, we introduce the Reconstructive Visual Head (RVH). This auxiliary module predicts the latent representation of the final transformed state, which corresponds to the final chronological image, from the shuffled candidate frames.

Let $v_{\text{final}} \in \mathcal{O}$ denote the candidate image corresponding to the final chronological state. We use its unpooled sequence embedding from the frozen visual encoder as the RVH supervision target:
\[
h_{\text{final}}
=
\text{Encoder}_{\text{vis}}(v_{\text{final}})
\in
\mathbb{R}^{N_{\text{img}} \times d},
\]
where $N_{\text{img}}$ is the number of visual tokens for each image.

The RVH module $g_\psi(\cdot)$ is a two-layer multilayer perceptron operating in parallel with the text decoder. Let $\mathbf{H}_{\text{opt}} \in \mathbb{R}^{M \times d}$ denote the hidden states of the visual tokens corresponding to the shuffled candidate images produced by $\text{Encoder}_{\text{txt}}(\cdot)$. The RVH predicts the final-state latent representation as
\[
\hat{h}_{\text{final}}
=
g_\psi(\mathbf{H}_{\text{opt}}),
\qquad
\hat{h}_{\text{final}}
\in
\mathbb{R}^{N_{\text{img}} \times d}.
\]

We train the RVH using a Mean Squared Error (MSE) loss together with the standard autoregressive cross-entropy loss for the target sequence $y^*$:
\begin{equation}
\begin{aligned}
    \mathcal{L}_{\text{SFT}}
    &=
    -\sum_{t=1}^{|y^*|}
    \log \pi_\theta(y^*_t \mid x, y^*_{<t}) \\
    &\quad
    +
    \lambda_{\text{RVH}}
    \cdot
    \text{MSE}
    \left(
    \hat{h}_{\text{final}},
    h_{\text{final}}
    \right).
\end{aligned}
\end{equation}

Here, $\lambda_{\text{RVH}}$ balances the autoregressive and latent reconstruction objectives. This supervision connects sequence prediction with the visual representation of the final state.

\subsection{Supervised Fine-Tuning with Region of Interest Attention Locating}

Resolving abstract visual transformations requires tracking localized dynamic changes. To prevent scattered attention and semantic noise, we propose a Region of Interest (ROI) Attention Locating module. Guided by dataset annotations, we prompt the model to generate a concise semantic locate cue enclosed by specific structural tokens before outputting the sequence. This isolates the key visual evidence.

To prevent spatial signal fragmentation, we condense the attention into a designated intermediate layer $\ell$. We extract the text-to-image self-attention weights at layer $\ell$ using the generated locate tokens as queries and the initial query image tokens as keys. Using the ground-truth bounding box $\mathcal{R}_{\text{tar}}$ of the transformation area, we optimize spatial alignment via an Attention Condensation loss:

\begin{equation}
\mathcal{L}_{AC} = -\log\left(s^{(\ell)}(\mathcal{R}_{\text{tar}})\right)
\end{equation}

where $s^{(\ell)}(\mathcal{R}_{\text{tar}})$ is the normalized mean-attention ratio within the bounding box:

\begin{equation}
s^{(\ell)}(\mathcal{R}_{\text{tar}}) = \frac{\frac{1}{|\mathcal{R}_{\text{tar}}|} \sum_{p \in \mathcal{R}_{\text{tar}}} \tilde{a}_p^{(\ell)}}{\frac{1}{|\mathcal{R}_{\text{img}}|} \sum_{p \in \mathcal{R}_{\text{img}}} \tilde{a}_p^{(\ell)}}
\end{equation}

Here, $\mathcal{R}_{\text{img}}$ is the set of all image tokens, and $\tilde{a}_p^{(\ell)}$ is the aggregated attention weight at position $p$ in layer $\ell$, averaged over all query tokens and attention heads. The overall objective is $\mathcal{L}_{\text{total}} = \mathcal{L}_{\text{SFT}} + \alpha \mathcal{L}_{AC}$.

A detailed methodological discussion on why we prioritize attention alignment over explicit bounding box coordinate regression is provided in \ref{appendix:attention_vs_bbox}.

\subsection{Reinforcement Learning with Implicit Process Grounding}

To mitigate compounding errors in long-horizon reasoning, we introduce a reinforcement learning stage based on Group Relative Policy Optimization (GRPO) \citep{shao2024deepseekmath}. We design an implicit process grounding mechanism that decouples the reward function into three granular components bounded within $[0, 1]$, evaluating the reasoning trajectory without human annotation.

\vspace{1mm}
\noindent\textbf{1. Final Outcome Reward ($R_{\text{out}}$).} 
We assign a sparse reward based on the exact match between the predicted sequence $y$ and the ground-truth sequence $y^*$: 
\begin{equation}
R_{\text{out}} = \begin{cases} 1, & \text{if } y = y^* \\ 0, & \text{otherwise.} \end{cases}
\end{equation}

\vspace{1mm}\noindent\textbf{2. Latent-Grounding Process Reward ($R_{\text{latent}}$).}
We evaluate the reasoning process at the sentence level to provide dense supervision. Let $K$ be the total number of reasoning steps (sentences) in $y$. At step $k$, we feed the contextualized hidden states of the visual tokens into the RVH to generate a latent spatial feature sequence $Z_k \in \mathbb{R}^{N_{\text{img}} \times d}$. Simultaneously, the visual encoder provides candidate visual features $\mathbf{V} = \{V_i\}_{i=1}^6$, where $V_i = \text{Encoder}_{\text{vis}}(o_i) \in \mathbb{R}^{N_{\text{img}} \times d}$ corresponds to the shuffled options. The process reward measures the maximum cosine similarity between the reasoning intent $Z_k$ and the available visual features $V_i$:
\begin{equation}
    R_{\text{latent}} = \frac{1}{K} \sum_{k=1}^{K} \max_{V_i \in \mathbf{V}} \left( \frac{\cos(Z_k, V_i) + 1}{2} \right),
\end{equation}
where $\cos(\cdot, \cdot)$ denotes the flattened cosine similarity. This anchors textual reasoning to underlying image features.

\vspace{1mm}\noindent\textbf{3. Unsupervised Visual Focus Reward ($R_{\text{focus}}$).}
To prevent attention collapse, we introduce an unsupervised focus reward based on internal self-attention maps. Let $\mathbf{a}_k \in \mathbb{R}^{N_{\text{img}}}$ be the normalized self-attention probability distribution over the $N_{\text{img}}$ visual tokens at step $k$. Using its Shannon entropy $H(\mathbf{a}_k) = -\sum_{i=1}^{N_{\text{img}}} a_{k,i} \log a_{k,i}$, we formulate a normalized negative-entropy reward:
\begin{equation}
    R_{\text{focus}} = \frac{1}{K} \sum_{k=1}^{K} \left( 1 - \frac{H(\mathbf{a}_k)}{\log N_{\text{img}}} \right).
\end{equation}

\vspace{1mm}\noindent\textbf{GRPO Optimization Objective.}
The total reward is $R_{\text{total}} = \omega_1 R_{\text{out}} + \omega_2 R_{\text{latent}} + \omega_3 R_{\text{focus}}$, where $\omega_1, \omega_2,$ and $\omega_3$ are balancing weights.

For each prompt $x \sim \mathcal{D}$, the old policy $\pi_{\text{old}}$ samples $G$ outputs $\{y_1, \dots, y_G\}$. We compute the group-relative advantage $\hat{A}_i = (R_{\text{total}}^{(i)} - \mu_R) / \sigma_R$, where $\mu_R$ and $\sigma_R$ are the mean and standard deviation of the group rewards. The policy $\pi_\theta$ is optimized using a clipped surrogate objective with a Kullback-Leibler penalty:

\begin{table*}[t]
\centering
\caption{Benchmarking results on \benchname{}. Overall In-Domain (ID) and Out-of-Domain (OOD) exact-match accuracy is reported alongside category-wise performance. Higher is better. \textbf{Bold}: best in group; \underline{underline}: second best.}
\resizebox{1.0\linewidth}{!}{
\begin{tabular}{l|c|c|ccccc|c|ccccc}
\toprule
& \multicolumn{1}{c|}{}
& \multicolumn{6}{c|}{\textbf{In-Domain by Category}}
& \multicolumn{6}{c}{\textbf{Out-of-Domain by Category}} \\
\cmidrule(lr){3-3}
\cmidrule(lr){4-8}
\cmidrule(lr){9-9}
\cmidrule(lr){10-14}
\textbf{Models}
& \textbf{Overall}
& \textbf{Avg.}
& \textbf{Flu.} & \textbf{Cry.} & \textbf{Vis.} & \textbf{Men.} & \textbf{Trans.}
& \textbf{Avg.}
& \textbf{Flu.} & \textbf{Cry.} & \textbf{Vis.} & \textbf{Men.} & \textbf{Trans.} \\
\midrule

Qwen 3.5 9B~\citep{qwen35blog}
& 14.2 & 16.8 & 10.8 & 13.3 & 20.0 & 40.0 & 12.7
& 11.6 & 13.3 & 20.0 & 5.0 & 12.0 & 10.0 \\

GLM-4.6V~\citep{vteam2026glm45vglm41vthinkingversatilemultimodal}
& 21.4 & 19.2 & 10.8 & 22.2 & 21.7 & 20.0 & 23.6
& 23.6 & 15.6 & 30.0 & 25.0 & 16.0 & 26.0 \\

GPT-5.4~\citep{singh2025openai}
& 33.8 & 28.0 & 23.1 & 31.1 & 25.0 & 32.0 & 32.7
& 39.6 & 20.0 & 47.5 & 30.0 & 40.0 & 49.0 \\

Gemini 3.1 Pro~\citep{google-deepmind-gemini2025}
& 41.8 & \underline{52.8} & \underline{32.3} & \underline{66.7} & \underline{66.7} & 60.0 & 47.3
& 30.8 & 42.2 & 62.5 & \underline{47.5} & 48.0 & 2.0 \\

GPT o3~\citep{gpto3}
& 46.0 & 42.8 & 26.2 & 60.0 & 45.0 & 64.0 & 36.4
& 49.2 & 31.1 & 62.5 & 45.0 & \underline{60.0} & 51.0 \\

Gemini 3.0 Flash~\citep{google-deepmind-gemini-flash2025}
& 46.6 & 47.2 & 30.8 & 51.1 & 53.3 & \textbf{80.0} & 41.8
& 46.0 & 31.1 & 37.5 & 45.0 & 48.0 & 56.0 \\

Qwen 3.5 397B~\citep{qwen35blog}
& 49.4 & 45.6 & 21.5 & 60.0 & 58.3 & 44.0 & \underline{49.1}
& 53.2 & 42.2 & \underline{67.5} & \underline{47.5} & 44.0 & 57.0 \\

Claude Opus 4.6~\citep{anthropic2025claudeopus46}
& \underline{55.8} & 50.8 & 30.8 & \underline{66.7} & 60.0 & 56.0 & \underline{49.1}
& \underline{60.8} & \underline{48.9} & \underline{67.5} & 42.5 & 56.0 & \underline{72.0} \\

\midrule
\rowcolor{line-blue}
\textbf{\modelname}
& \textbf{73.2} & \textbf{74.8} & \textbf{72.3} & \textbf{77.8} & \textbf{75.0} & \underline{68.0} & \textbf{78.2}
& \textbf{71.6} & \textbf{66.7} & \textbf{75.0} & \textbf{72.5} & \textbf{64.0} & \textbf{74.0} \\
\bottomrule
\end{tabular}
    }
\label{tab:ordering_results}
\end{table*}

\begin{equation}
\begin{aligned}
    \mathcal{L}_{\text{GRPO}}(\theta) &= \mathbb{E}_{x \sim \mathcal{D},\, y \sim \pi_{\text{old}}}
    \Bigg[ \frac{1}{G} \sum_{i=1}^G \Bigg( \\
    &\quad \min \Big( r_i \hat{A}_i,\;
    \text{clip}(r_i,\, 1\!-\!\epsilon,\, 1\!+\!\epsilon) \hat{A}_i \Big) 
    \\
    &\quad - \beta \log \frac{\pi_\theta(y_i|x)}{\pi_{\text{ref}}(y_i|x)}
    \Bigg) \Bigg],
\end{aligned}
\end{equation}
where $r_i = \frac{\pi_\theta(y_i|x)}{\pi_{\text{old}}(y_i|x)}$.
\section{Experiments}

\subsection{Experimental Setup}

\textbf{Implementation Details. }We evaluate \modelname{} on the Vbvr-VQA benchmark to test fine-grained visual understanding and logical reasoning capabilities in visual temporal transformations. More implementation details will be in Appendix \ref{appendix:implement}. We report the Exact Match accuracy as the primary evaluation metric. A prediction is considered correct if and only if the entire generated sequence of the six candidate frames exactly matches the ground-truth chronological order.

\textbf{Baselines. }We benchmark \modelname{} against state-of-the-art open-source and proprietary vision-language models. The open-source baselines, selected for their strong spatial reasoning and recognition capabilities, include Qwen 3.5 397B~\citep{qwen35blog}, GLM-4.6V~\citep{vteam2026glm45vglm41vthinkingversatilemultimodal}, and Qwen 3.5 9B~\citep{qwen35blog}. To establish a reference for standard zero-shot capabilities, we also evaluate widely used commercial models, specifically Gemini 3.1 Pro~\citep{google-deepmind-gemini2025}, Gemini 3.0 Flash~\citep{google-deepmind-gemini-flash2025}, Claude Opus 4.6~\citep{anthropic2025claudeopus46}, GPT o3~\citep{gpto3}, and GPT-5.4~\citep{singh2025openai}.

\subsection{Main Results}

\begin{figure*}[!t]
    \centering
    \includegraphics[width=1.0\linewidth]{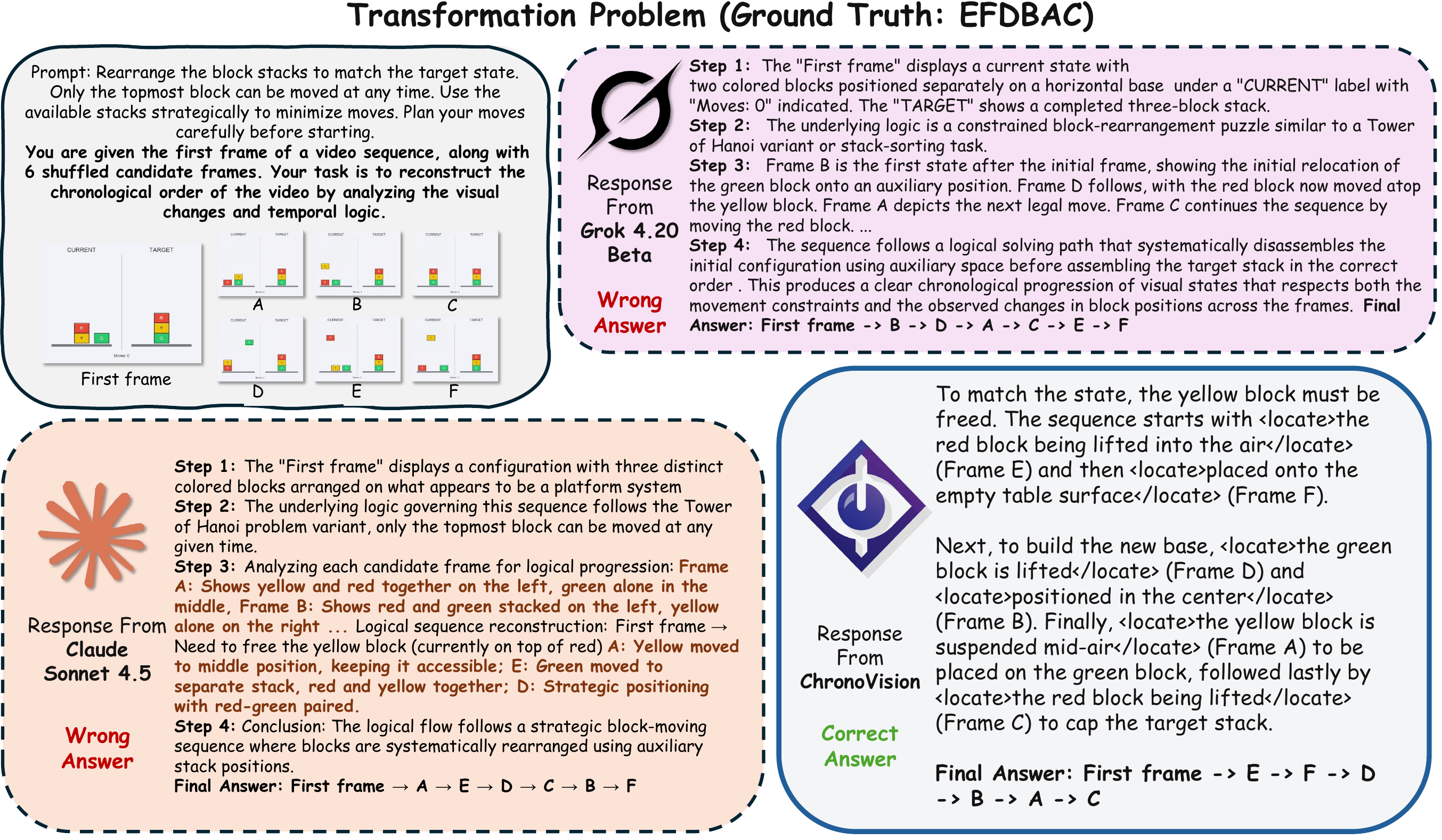}
    \caption{Qualitative comparison of reasoning chains on a transformation planning problem from Vbvr-VQA. The task requires reconstructing the correct chronological order of six shuffled candidate frames by inferring valid intermediate moves under the top-block-only constraint.}
    \label{fig:example1}
\end{figure*}

\textbf{Performance on Vbvr-VQA.} As shown in Table ~\ref{tab:ordering_results}, ChronoVision establishes a new state-of-the-art on the Vbvr-VQA benchmark, achieving 74.8\% accuracy on the In-Domain split, 71.6\% on the Out-of-Domain split, and 73.2\% overall. It surpasses the strongest proprietary baseline, Claude Opus 4.6 (50.8\% ID, 60.8\% OOD). Our 9B parameter framework successfully outperforms models with significantly larger parameter scales, including Qwen 3.5 397B and GPT o3. The Figure \ref{fig:example1} shows the visualization of qualitative comparison of reasoning chains on a transformation planning problem from Vbvr-VQA. And more examples will be in Appendix \ref{appendix:example_vis}

\textbf{General Vision-Language Capabilities.} We evaluate ChronoVision on 7 standard multimodal benchmarks. As presented in Table \ref{tab:general_ability}, ChronoVision maintains performance strictly comparable to the vanilla Qwen 3.5 9B across all general tasks (e.g., 78.8 vs. 78.4 on MMMU, 85.9 vs. 85.7 on MathVista), confirming that our attention locating and reconstructive objectives enhance spatial-temporal reasoning without degrading general visual-language understanding.

\begin{table*}[t!]
  \centering
  \caption{\textbf{General Vision-Language Understanding Results.} Best performance in \textbf{bold}.}
  \label{tab:general_ability}
  \resizebox{\textwidth}{!}{%
  \begin{NiceTabular}{l c c c c c c c}
    \toprule[1pt]
    \textbf{Models} & \textbf{\textsc{MMMU}} & \textbf{\textsc{MMBench}} & \textbf{\textsc{MMStar}} & \textbf{\textsc{MathVista}} & \textbf{\textsc{AI2D}} & \textbf{\textsc{HallusionBench}} & \textbf{\textsc{BabyVision}} \\
    \midrule
    
    Qwen3.5 9B~\citep{qwen35blog} & 78.4 & \textbf{90.1} & \textbf{79.7} & 85.7 & \textbf{90.2} & 69.3 & 25.8 \\
    
    \rowcolor{line-blue} \textbf{\modelname{}} & \textbf{78.8} & 88.2 & 78.5 & \textbf{85.9} & \textbf{90.2} & \textbf{69.6} & \textbf{27.3} \\
    \bottomrule
  \end{NiceTabular}}
\end{table*}

\subsection{Ablation Studies}
\begin{table*}[t]

\centering
\scriptsize
\setlength{\tabcolsep}{3pt}
\caption{Ablation study results on training processes and rewards (Accuracy \%). Overall In-Domain and Out-of-Domain accuracy metrics are reported alongside category-wise performance. Higher is better. \textbf{Bold}: best in group; \underline{underline}: second best.}

\label{tab:combined_ablation_results}

\resizebox{1.0\linewidth}{!}{
\begin{tabular}{l|c|c|ccccc|c|ccccc}
\toprule
& \multicolumn{1}{c|}{}
& \multicolumn{6}{c|}{\textbf{In-Domain by Category}}
& \multicolumn{6}{c}{\textbf{Out-of-Domain by Category}} \\
\cmidrule(lr){3-3}
\cmidrule(lr){4-8}
\cmidrule(lr){9-9}
\cmidrule(lr){10-14}
\textbf{Models}
& \textbf{Overall}
& \textbf{Avg.}
& \textbf{Flu.} & \textbf{Cry.} & \textbf{Vis.} & \textbf{Men.} & \textbf{Trans.}
& \textbf{Avg.}
& \textbf{Flu.} & \textbf{Cry.} & \textbf{Vis.} & \textbf{Men.} & \textbf{Trans.} \\
\midrule

\rowcolor{line-blue}\textbf{Training Processes} & & & & & & & & & & & & & \\

SFT w/o head
& 66.0 & 66.8 & 66.2 & 73.3 & 66.7 & 56.0 & 67.3
& 65.2 & 62.2 & 70.0 & 65.0 & 52.0 & 68.0 \\

SFT only
& 69.0 & 70.0 & \underline{69.2} & \underline{75.6} & 68.3 & 60.0 & 72.7
& 68.0 & \underline{64.4} & \underline{72.5} & 67.5 & \underline{56.0} & 71.0 \\

SFT with ROI
& \underline{70.2} & \underline{71.6} & \underline{69.2} & \underline{75.6} & \underline{71.7} & \underline{64.0} & \underline{74.5}
& \underline{68.8} & \underline{64.4} & \underline{72.5} & \underline{70.0} & \underline{56.0} & \underline{72.0} \\

SFT with ROI + RL
& \textbf{73.2} & \textbf{74.8} & \textbf{72.3} & \textbf{77.8} & \textbf{75.0} & \textbf{68.0} & \textbf{78.2}
& \textbf{71.6} & \textbf{66.7} & \textbf{75.0} & \textbf{72.5} & \textbf{64.0} & \textbf{74.0} \\

\midrule
\rowcolor{line-blue}\textbf{Reward Components} & & & & & & & & & & & & & \\

w/o Answer Reward
& 69.8 & 71.2 & 69.2 & \underline{75.6} & 70.0 & \underline{64.0} & 74.5
& 68.4 & \underline{64.4} & \underline{72.5} & \underline{70.0} & 56.0 & 71.0 \\

w/o Latent-Grounding Reward
& 71.0 & 72.8 & \underline{70.8} & \underline{75.6} & 71.7 & \textbf{68.0} & \underline{76.4}
& 69.2 & \underline{64.4} & \underline{72.5} & \underline{70.0} & \underline{60.0} & 72.0 \\

w/o ROI Reward
& \underline{71.8} & \underline{74.0} & \underline{70.8} & \textbf{77.8} & \underline{73.3} & \textbf{68.0} & \textbf{78.2}
& \underline{69.6} & \underline{64.4} & \underline{72.5} & \underline{70.0} & \underline{60.0} & \underline{73.0} \\

Full Reward
& \textbf{73.2} & \textbf{74.8} & \textbf{72.3} & \textbf{77.8} & \textbf{75.0} & \textbf{68.0} & \textbf{78.2}
& \textbf{71.6} & \textbf{66.7} & \textbf{75.0} & \textbf{72.5} & \textbf{64.0} & \textbf{74.0} \\

\bottomrule
\end{tabular}
}
\end{table*}
We conduct ablation studies to isolate the contributions of our architectural components and RL reward designs, as detailed in Table \ref{tab:combined_ablation_results}.  We provide additional, detailed ablation studies in Appendix \ref{appendix:ablation} covering partial match evaluation, the evolution of latent representations, the impact of chain-of-thought reasoning, prefix sensitivity, intermediate state perturbations, latent sequence interventions, linear probing, and comparisons with related temporal reasoning methods.

\textbf{Effectiveness of Training Processes.} 
Starting from the baseline model without the Reconstructive Visual Head (\textit{SFT w/o head}), adding the RVH (\textit{SFT only}) yields a direct +3.2\% ID and +2.8\% OOD improvement. This confirms that explicitly predicting latent visual states effectively grounds sequence ordering. Integrating the ROI Attention Locating module (\textit{SFT with ROI}) further improves overall accuracy by +1.2\% (from 69.0\% to 70.2\%), proving its ability to filter spatial semantic noise. Finally, applying the GRPO reinforcement learning stage (\textit{SFT with ROI + RL}) provides another substantial boost (+3.2\% ID, +2.8\% OOD), demonstrating that RL effectively suppresses error accumulation during long-horizon temporal reasoning.

\textbf{Ablation on Reward Components.} 
Within the RL stage, we ablate the composite reward function to verify that all components are indispensable. Removing the sparse \textit{Answer Reward} (outcome reward $R_{\text{out}}$) causes the most severe performance drop (-3.4\% Overall). Crucially, removing the dense intermediate signals also degrades performance: omitting the \textit{Latent-Grounding Process Reward} (latent alignment process reward $R_{\text{latent}}$) leads to a -2.2\% overall decline, while removing the \textit{ROI Reward} (unsupervised visual focus reward $R_{\text{focus}}$) results in a -1.4\% overall drop. This verifies that both latent trajectory guidance and forced spatial attention are critical for maximizing reasoning accuracy. \looseness=-1

\subsection{Out-of-Domain Evaluation on Physical Dynamics}

\begin{figure*}[t]
    \centering
    \includegraphics[width=1.0\linewidth]{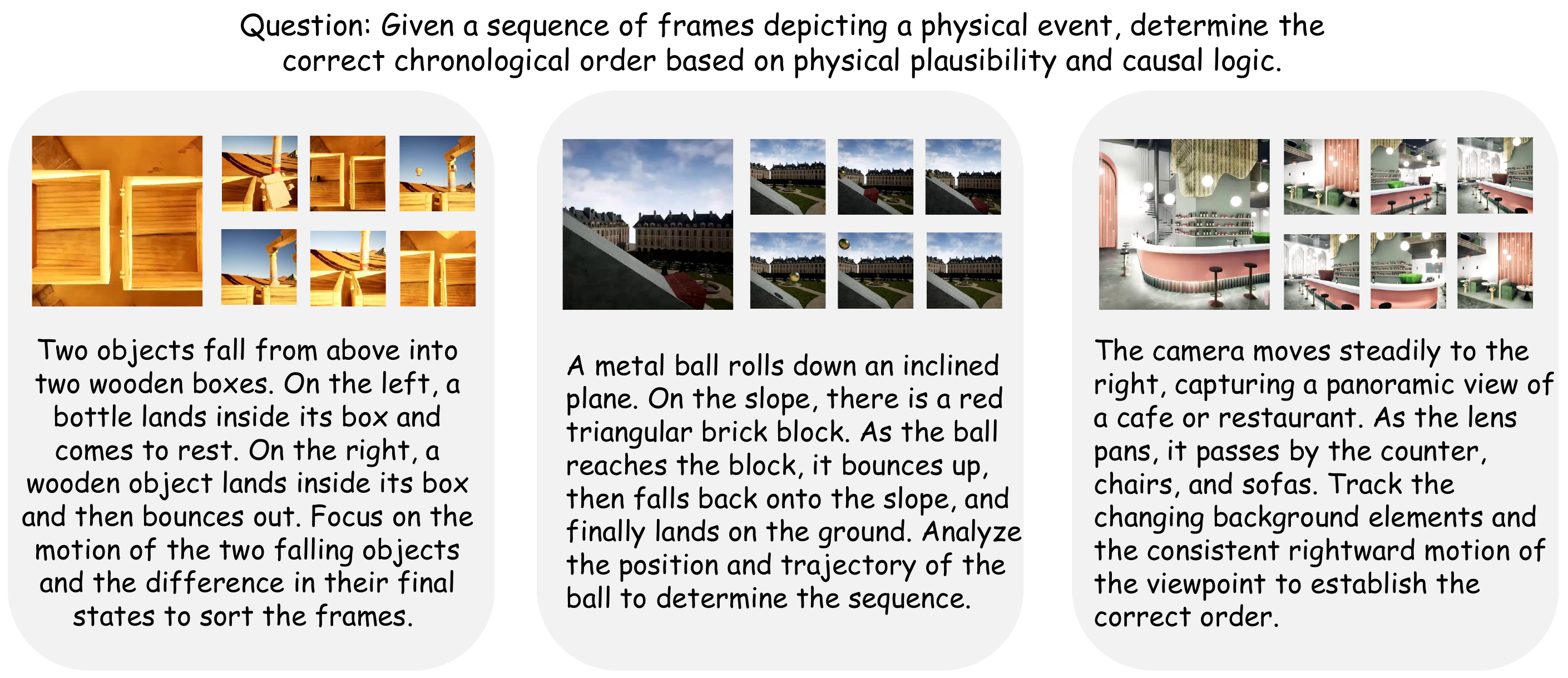} 
    \caption{Examples of out-of-domain physical reasoning. We visualize three distinct real-world scenarios where the model must reconstruct the chronological order based on physical plausibility and causal logic. \textbf{Left:} Two objects fall into boxes exhibiting different kinetic behaviors (one resting, one bouncing), requiring the model to infer collision outcomes. \textbf{Middle:} A metal ball rolls down a slope, hits a stationary block, and bounces, testing intuitive understanding of momentum and trajectory. \textbf{Right:} A camera pans steadily across a café, challenging the model to track spatial consistency and continuous viewpoint shifts in a complex real-world environment.}
    \label{fig:intphys2_examples}
\end{figure*}

\begin{table}[t!]
  \centering
  \caption{\textbf{Generalization on IntPhys 2.} Out-of-domain performance evaluating cognitive understanding of universal physical laws. Accuracy (\%) is reported. Best performance in \textbf{bold}.}
  \label{tab/intphys2_results}
  \resizebox{\columnwidth}{!}{%
    \begin{NiceTabular}{l c c c c}
      \toprule[1pt]
      \textbf{Models} & \textbf{Easy} & \textbf{Medium} & \textbf{Hard} & \textbf{Overall} \\
      \midrule
      
      Qwen 3.5 9B~\citep{qwen35blog} & 51.0 & 48.5 & 47.6 & 48.5 \\
      
      \rowcolor{line-blue} \textbf{\modelname{}} & \textbf{62.5} & \textbf{54.8} & \textbf{53.0} & \textbf{55.0} \\
      \bottomrule
    \end{NiceTabular}%
  }
\end{table}

While our in-domain evaluation focused on abstract cognitive reasoning (e.g., spatial puzzles and block transformations), we further evaluate our model's generalization capabilities on real-world physical dynamics using the highly challenging IntPhys 2 benchmark \citep{bordes2025intphys}. Rather than abstract shapes, this benchmark requires models to sequence frames depicting realistic physical events based on causal logic and physical plausibility (e.g., gravity, collisions, and momentum). Examples of these realistic scenarios are illustrated in Figure~\ref{fig:intphys2_examples}.

Evaluating intuitive physics is notoriously difficult for current Vision-Language Models (VLMs), with most baselines hovering near random chance (approximately 50\%). However, as detailed in Table~\ref{tab/intphys2_results}, ChronoVision achieves consistent improvements across the Easy, Medium, and Hard subsets of IntPhys 2. 

Overall, our model attains an accuracy of 55.0\%, yielding a +6.5\% absolute gain over the strong base model, Qwen 3.5 9B (48.5\%). The improvement is particularly pronounced on the Easy subset, where ChronoVision reaches 62.5\%, outperforming the baseline by +11.5\%. These substantial OOD gains confirm that our framework enables the model to internalize generalized physical laws and spatial-temporal continuity.

We also evaluate ChronoVision on two real-world video reasoning benchmarks, Video-Holmes and LongVideo-Reason, where it obtains overall scores of 45.89 and 74.7, respectively. The complete results are reported in Appendix~\ref{sec:real_world_video}.
\section{Related Works}
Our work is closely related to VLMs for Vision Reasoning, Visual Cognition, ROI Selection and Cropping and Reinforcement learning. A comprehensive discussion
is provided in Appendix \ref{sec:appendix_related_works}

\section{Conclusion}

This paper introduces ChronoVision, a framework designed to resolve the text bottleneck in long horizon abstract visual reasoning. By integrating a Reconstructive Visual Head and a Region of Interest Attention module during supervised fine tuning, the model acquires implicit visual reasoning capabilities and fine grained spatial focus. We further propose a reinforcement learning stage utilizing a composite reward function to mitigate compounding errors in textual chains of thought. Evaluated on our proposed Vbvr-VQA benchmark and the cross domain IntPhys2 dataset, ChronoVision achieves state of the art performance against open source and proprietary models. Future research will explore integrating generative models to explicitly visualize internal visual chains of thought.

\section*{Limitation}

While ChronoVision demonstrates strong temporal reasoning and visual cognition capabilities on Vbvr-VQA and IntPhys2, the current study primarily focuses on validating the effectiveness of latent sequence reconstruction and process grounding within a moderate-scale 9B MLLM setting. Future work could further investigate scalability to larger multimodal backbones and larger and more diverse training corpora. Besides, the current framework relies on dense auxiliary supervision, including semantic locating cues and spatial bounding-box annotations, to guide the ROI Attention Locating module. Future research may explore weaker-supervision or annotation-free alternatives to improve scalability and adaptability across broader video reasoning domains.

\bibliography{custom}

\newpage
\appendix
\section{Additional information of the Dataset}
\label{appendix:dataset_vis}

\begin{figure*}[t]
    \centering
    \includegraphics[width=\linewidth]{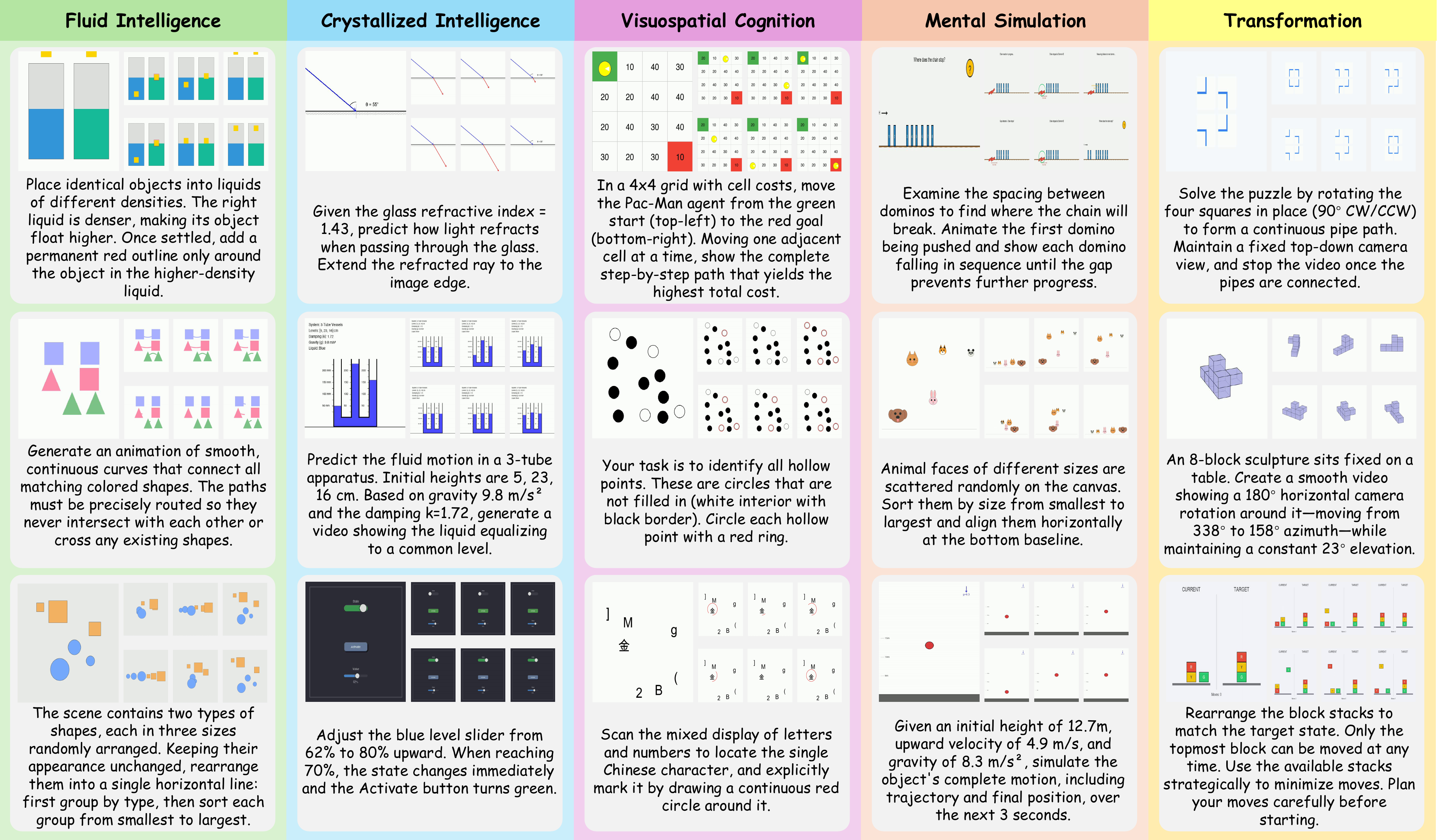} 
    \caption{More Examples of Vbvr-VQA dataset}
    \label{fig:appendix_dataset}
\end{figure*}

\subsection{Statistics of the Dataset}

Our Vbvr-VQA dataset is constructed by combining and processing the VBVR-Dataset and VBVR-Bench-Data from the Very Big Video Reasoning (VBVR) Suite \citep{wang2026bigvideoreasoningsuite}. Specifically, the VBVR-Dataset serves as the large-scale video reasoning training resource for the VBVR Suite. The original dataset is massive, containing 1,000,000 video clips generated by 100 distinct reasoning task generators (10,000 samples per generator). This results in approximately 5,000,000 individual files with a total size of roughly 370 GB. Furthermore, the VBVR-Bench-Data contains 500 test samples covering 100 tasks. Among these, 50 tasks are used for in-domain testing, and the remaining 50 are used for out-of-domain testing. 

The data splits for the training and testing sets of Vbvr-VQA strictly follow the strategy of the original VBVR Suite. We use the frame ordering tasks processed from the VBVR-Dataset to train the ChronoVision framework. Subsequently, we evaluate the model on the tasks processed from the VBVR-Bench-Data. To ensure a rigorous evaluation, the training and testing tasks are generated using different random seeds.

The structure of our Vbvr-VQA is organized as follows.

\textbf{Training Set}: A total of 1,000,000 samples are used for the Supervised Fine-Tuning (SFT) and Reinforcement Learning (RL) stages of ChronoVision.

\textbf{Testing Set}: A total of 500 samples are used for the final benchmark evaluation. To rigorously test the model's generalization capabilities, this set is strictly divided into two parts: an \textbf{In-Domain Test Split} with 250 samples, where the task types and physical scenarios have appeared in the training set, and a \textbf{Zero-Shot Out-of-Domain (OOD) Test Split} with 250 samples. The OOD split contains completely new task generators and extreme physical scenarios that the model has never seen during the training phase, and is specifically designed to evaluate the generalized understanding of universal physical laws.

While the test set comprises 500 samples, this scale is strictly sufficient to demonstrate robust reasoning capabilities due to the mathematical formulation of the task and the depth of the physical scenarios. Mathematically, ordering six unordered candidate frames results in $6! = 720$ possible permutations. Consequently, the probability of correctly guessing the exact chronological order by chance is exceptionally low ($1/720 \approx 0.14\%$). Achieving an accuracy exceeding 70 percent across 500 instances, particularly on the 250 out-of-domain samples, provides strong statistical significance. This proves that the model executes genuine temporal reasoning rather than relying on random selection or superficial statistical shortcuts.

Beyond statistical significance, the evaluation depth of each individual sample far exceeds standard visual question answering tasks. As illustrated in Figure \ref{fig:appendix_dataset}, the test set covers 100 distinct reasoning task generators spanning complex cognitive domains such as fluid dynamics, spatial rotation, and kinematics. Each instance requires the model to perform multi-step mental simulation to reconstruct the visual evolution. The out-of-domain split introduces entirely unseen physical environments and extreme scenarios. This density of physical complexity justifies the test set scale, as the evaluation focuses on the rigorous depth of cognitive simulation rather than simple pattern recognition over massive data volumes.

\subsection{Visual Examples of Dataset}

To further illustrate the diversity of our Vbvr-VQA benchmark, we provide additional visual examples. As shown in Figure~\ref{fig:MainBaby_fig} (in the main text), our dataset covers five distinct cognitive categories: Fluid Intelligence, Crystallized Intelligence, Mental Simulation, Visuospatial Cognition, and Transformation. In Figure~\ref{fig:appendix_dataset}, we present another set of examples. These samples intuitively demonstrate the broad coverage, physical complexity, and high level of challenge in the tasks.

As mentioned in the main text, we reformulate the continuous video streams into a strict image ordering task. We uniformly segment the video along the temporal axis into six equal-length intervals and extract the final frame of each interval. These six discrete frames precisely capture the dynamic temporal evolution of the scene. Subsequently, we randomly shuffle these six images (labeled A through F) to serve as the candidate options for the model.

\subsection{Annotation Details of Dataset}

The text descriptions (prompts) are constructed by concatenating task instructions (requiring the model to order the images, such as "Please determine the correct chronological order of the following images based on physical plausibility and causal logic") with the original video descriptions provided by the VBVR dataset (which serve as prompts introducing the temporal logic). We use these combined prompts as the reasoning queries for the Multimodal Large Language Model (MLLM). During the inference process, the model receives the initial frame, the text prompt, and the six shuffled candidate frames. This evaluates its ability to reconstruct the correct chronological order based on the given instructions.

\begin{figure*}[!t]
    \centering
    \includegraphics[width=1.0\linewidth]{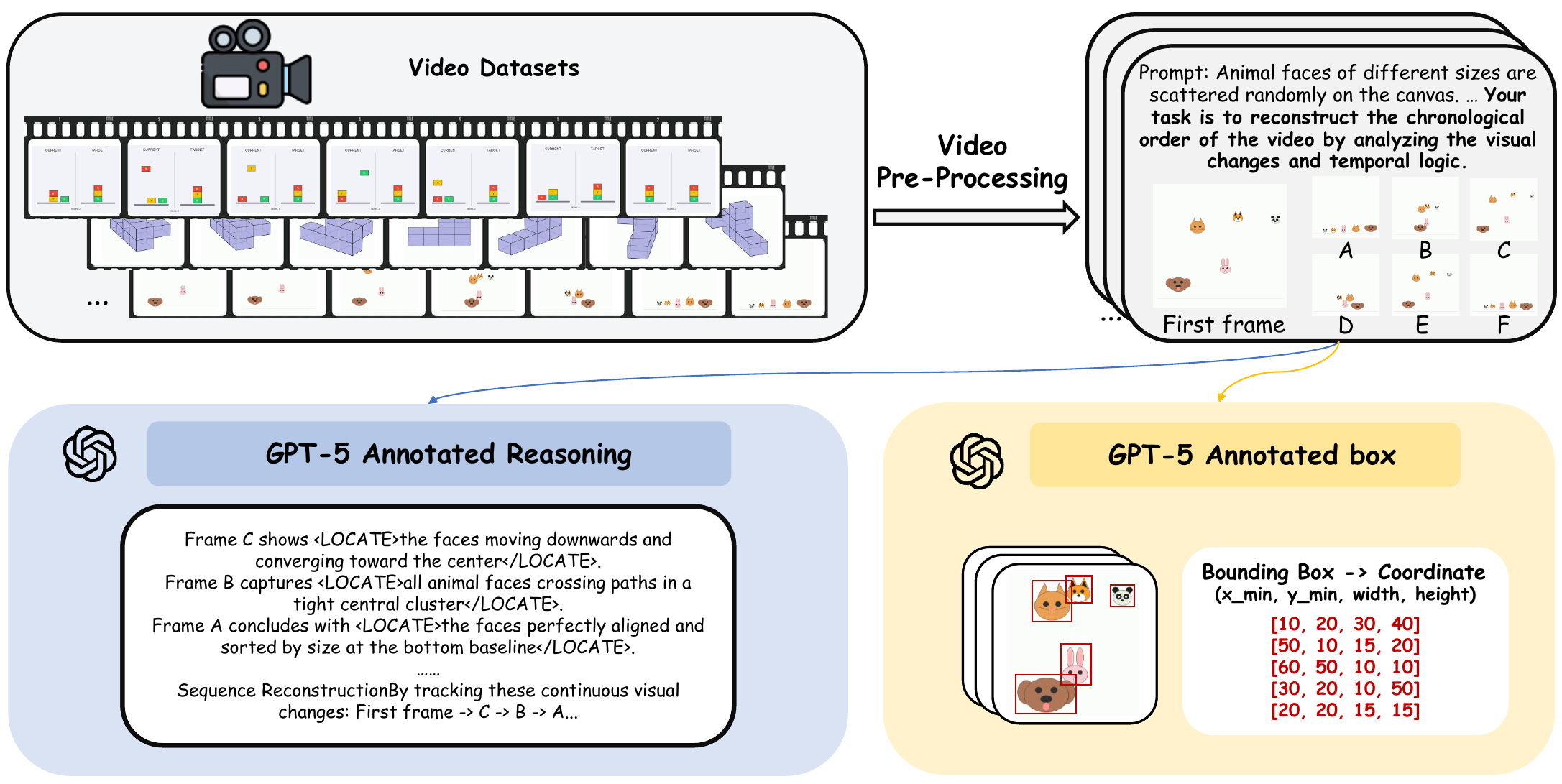}
    \caption{Dataset Processing Pipeline. }
    \label{fig:dataset_pipeline}
\end{figure*}

Additionally, we use GPT-5 to generate dense spatio-temporal annotations. These annotations include not only precise bounding box coordinates for locating state change regions but also verb-centric semantic cues encapsulated in the \texttt{<LOCATE>...</LOCATE>} format. This provides high-quality supervision signals for our ROI Attention Locating module. We show our dataset pipeline in Figure \ref{fig:dataset_pipeline} and related prompts in Figure \ref{fig:appx_prompt}.

\subsection{Data Quality Assurance and Human Validation}

A legitimate concern when utilizing data annotated by large language models is the potential risk of distilling hallucinated logic or erroneous physical assumptions. To prevent this, we strictly decouple the sequence determination from the text generation process. The chronological order of the candidate frames is an absolute ground truth, intrinsically defined by the continuous temporal axis of the original source video. GPT-5 is not tasked with inferring or guessing this sequence. Instead, the model is explicitly conditioned on the fixed ground-truth order. Its sole objective is to synthesize intermediate reasoning steps and extract spatial bounding box coordinates that explain the predetermined visual evolution. 

To guarantee the validity of the evaluation benchmark, human annotators manually reviewed all 500 instances in the testing set. This review process verified that the generated temporal logic and semantic visual cues accurately reflect the actual physical events without ambiguity. Furthermore, to evaluate the reliability of the training data, we conducted a quality check by randomly sampling 500 instances from the training set. The manual inspection confirmed that the generated rationales logically align with the visual transformations. This rigorous verification confirms that the dataset provides physically sound supervision rather than flawed synthetic logic.

For further details on the ethics statement and reproducibility, please refer to Appendix \ref{appendix: G}.

\section{More Implementation Details}
\label{appendix:implement}

In this section, we provide extended details regarding the training paradigm, hyperparameters, and the specific mechanisms of the Attention Condensation module that were omitted from the main text due to space constraints. Note that the Reconstructive Visual Head (RVH) operates in parallel to these mechanisms, as detailed in Section \ref{method_subsec:RVH}.

\subsection{Training Paradigm and Hyperparameters}
All training and inference processes are conducted on 8 NVIDIA H100 GPUs, utilizing DeepSpeed ZeRO-3 optimization to manage memory efficiency. The overall training pipeline is divided into two sequential stages: the Supervised Fine-Tuning (SFT) stage and the Reinforcement Learning (RL) stage.

\textbf{SFT Stage.} During the SFT stage, we optimize the model using the AdamW optimizer with a peak learning rate of $2 \times 10^{-5}$, a global batch size of 128, and a weight decay of 0.05. The model is trained for 3 epochs using a cosine learning rate decay schedule with a 3\% linear warmup. The overall objective is 
\begin{equation}
\begin{split}
\mathcal{L}_{\text{total}} = & -\sum_{t=1}^{|Y|} \log q_\theta(Y_t \mid X, Y_{<t}) \\
& + \beta \cdot \text{MSE}(\hat{h}_{\text{final}}, h_{\text{final}}) + \alpha \mathcal{L}_{AC}
\end{split}
\end{equation}
We strictly decouple the loss balancing: we set $\beta = 0.1$ for the RVH term and $\alpha = 0.1$ for the Attention Condensation loss. 

\textbf{RL Stage.} For the subsequent RL stage utilizing Group Relative Policy Optimization (GRPO), we reduce the learning rate to $1 \times 10^{-6}$ to ensure policy stability. For each input prompt, we sample a group size of $G = 8$ responses. The clipping parameter is set to $\epsilon = 0.2$, and the Kullback-Leibler (KL) divergence penalty coefficient is fixed at $0.01$. The reward balancing weights are empirically set to $\omega_1 = 1.0$ for the outcome reward, $\omega_2 = 0.5$ for the latent-grounding reward, and $\omega_3 = 0.1$ for the unsupervised focus reward.

\subsection{Justification for Attention Alignment over Bounding Box Regression}
\label{appendix:attention_vs_bbox}

While ground-truth bounding box coordinates are available in the Vbvr-VQA dataset, we deliberately utilize them to supervise internal attention distributions rather than optimizing the model to directly output explicit coordinate values (e.g., via an Intersection over Union loss). This design choice is driven by the specific requirements of multi-step visual reasoning and the nature of physical simulation.

First, our core objective is to facilitate latent visual imagery. Explicitly predicting four discrete numerical coordinates compresses rich visual and spatial data into an extreme information bottleneck. By contrast, attention alignment preserves the high-dimensional, dense visual feature flow. These dense features are continuously processed by the Reconstructive Visual Head (RVH) to predict subsequent spatial states, maintaining feature integrity within the latent space.

Second, the Vbvr-VQA benchmark encompasses highly complex and non-rigid physical transformations, such as fluid dynamics, continuous smooth curves, and distributed objects. A rigid rectangular bounding box cannot accurately encapsulate irregular shapes or multiple scattered targets undergoing simultaneous state changes. Attention maps operate as pixel- or patch-level probability distributions, naturally accommodating arbitrary shapes and distributed regions without the structural constraints of explicit bounding boxes.

Finally, existing literature indicates that directly regressing continuous coordinate values using standard text-based Transformer architectures is often unstable and inefficient, frequently requiring specialized coordinate tokenizers. Supervising the self-attention mechanism via distribution alignment aligns directly with the fundamental architectural logic of multimodal large language models, ensuring stable spatial grounding.

\subsection{Selection of the Designated Attention Layer}
A core component of our methodology is condensing fragmented visual attention into a single designated intermediate layer $\ell$. The selection of this layer $\ell$ is established prior to the fine-tuning phase via a heuristic approach, ensuring it does not overfit to specific downstream QA tasks.

To ensure generalizability, we evaluate the base (unfine-tuned) backbone on a held-out visual reasoning validation set. We extract the text-to-image self-attention weights across all layers and select the single layer that naturally exhibits the highest average attention concentration on the target bounding boxes $\mathcal{R}_{\text{tar}}$. Empirical observations demonstrate that the optimal layer $\ell$ consistently falls in the intermediate region of the Transformer blocks across different MLLM backbones. We freeze this layer selection for all subsequent training phases.

We explicitly opt for a \textit{single} designated layer rather than aggregating multiple layers. Our preliminary experiments indicate that multi-layer aggregation dilutes the localization signal and introduces semantic noise from adjacent layers, whereas a single carefully selected intermediate layer provides the sharpest grounding interface for ROI extraction.

\section{Additional Analyses and Ablation Studies}
\label{appendix:ablation}

\subsection{Ablation Study on Evaluation with Partial Match Metrics}

We select Exact Match as the primary metric because physical causality exhibits an all-or-nothing characteristic. In real-world continuous dynamics, such as fluid motion or object trajectories, reversing the chronological position of even two frames implies a fundamental break in the causal chain. Therefore, Exact Match serves as the strictest standard to verify whether the model fully comprehends the complete physical progression, rather than merely approximating visual states.

While Exact Match imposes a strict physical constraint, it assigns a score of zero to predictions with minor localized errors. To provide a more granular assessment of the models' ability to capture the general temporal trend, we evaluate the baseline models using Kendall's $\tau$ rank correlation coefficient as a partial match metric. 

Kendall's $\tau$ measures the proportion of image pairs whose relative sequence is consistent between the predicted and ground-truth orders. For our task involving $N = 6$ frames, the total number of distinct frame pairs is $\binom{6}{2} = 15$. Let $C$ denote the number of concordant pairs (where the predicted relative order matches the ground truth) and $D$ denote the number of discordant pairs. Since there are no ties in our strict ordering format, $C + D = 15$. The metric is defined as:
\begin{equation}
\tau = \frac{C - D}{C + D}
\end{equation}
This metric effectively isolates minor errors. Table \ref{tab:kendall_tau} presents the Kendall's $\tau$ scores for ChronoVision and four prominent baselines across the ten cognitive categories. Even under this relaxed evaluation criterion, ChronoVision consistently demonstrates superior temporal alignment compared to the strong commercial and open-source models.

\begin{table*}[h]
\centering
\small
\setlength{\tabcolsep}{3pt}
\begin{tabular}{l|ccccc|ccccc}
\toprule
& \multicolumn{5}{c|}{\textbf{In-Domain by Category}} & \multicolumn{5}{c}{\textbf{Out-of-Domain by Category}} \\
\textbf{Models} & \textbf{Flu.} & \textbf{Cry.} & \textbf{Vis.} & \textbf{Men.} & \textbf{Trans.} & \textbf{Flu.} & \textbf{Cry.} & \textbf{Vis.} & \textbf{Men.} & \textbf{Trans.} \\
\midrule
Qwen 3.5 9B  & 29.5 & 31.4 & 36.8 & 51.2 & 30.9 & 31.6 & 36.2 & 25.1 & 30.7 & 29.4 \\
Gemini 3 Flash & 55.6&68.0&69.6&84.4&62.3&55.9&59.4&64.0&65.8&71.4 \\
Gemini 3.1 Pro & 56.1&78.4&78.8&73.8&65.8&63.0&76.2&66.6&66.9&24.8 \\
Claude Opus 4.6 & 58.8&80.1&76.2&74.1&69.9&70.1&81.0&66.4&74.2&83.5 \\
\midrule
\rowcolor{line-blue} \textbf{ChronoVision} & \textbf{88.4} & \textbf{90.7} & \textbf{89.5} & \textbf{87.8} & \textbf{91.0} & \textbf{86.9} & \textbf{89.5} & \textbf{88.8} & \textbf{85.5} & \textbf{89.3} \\
\bottomrule
\end{tabular}
\caption{Partial match evaluation using Kendall's $\tau$. We report the metric across all In-Domain and Out-of-Domain categories. Overall and average metrics are omitted to focus on specific cognitive dimensions.}
\label{tab:kendall_tau}
\end{table*}

\subsection{Ablation Study on Latent Sequence Evolution}

\label{appendix:evolution}
To validate the effectiveness of the RVH and explore the evolution if the latent sequence generated by it, we track the trajectories of the predicted latent sequence during the reasoning process for correct and incorrect predicted image orderings. Specifically, we measure the MSE between the predicted latent sequence $\hat{h}_{\mathrm{final}}$ generated by the RVH at various steps of the reasoning chain and the ground-truth sequence embedding of the final state $h_{\mathrm{final}}$ extracted by the frozen visual encoder. The results are averaged over 10 sampled reasoning trajectories. As shown in Table~\ref{tab:exp_evolution}, for correctly ordered trajectories, the MSE decreases as the reasoning step increases, providing evidence that the RVH successfully guides the model to mentally simulate and converge toward the correct final ordering sequence step by step. For incorrect trajectories, the MSE fails to converge to the ground-truth latent sequence. \looseness=-1

\begin{table}[h]
    \centering
    \caption{Latent sequence evolution across reasoning steps. We report the MSE between the predicted final-state latent representation and the ground-truth final-state representation for correct and incorrect ordering trajectories.}
    \label{tab:exp_evolution}
    \setlength{\tabcolsep}{6pt}
    \renewcommand\arraystretch{1.15}
    \resizebox{\linewidth}{!}{%
      \begin{tabular}{lccccc}
        \toprule
        \textbf{Final Answer} & \textbf{Step 0} & \textbf{Step 10} & \textbf{Step 20} & \textbf{Step 30} & \textbf{Step 40} \\
        \midrule
        Correct Answer & 0.512 & 0.364 & 0.245 & 0.182 & 0.146 \\
        Incorrect Answer & 0.508 & 0.456 & 0.412 & 0.435 & 0.428 \\ 
        \bottomrule
      \end{tabular}
    }
\end{table}

\subsection{Ablation Study on Chain-of-Thought}
\label{appendix:cot}

 To validate that the latent sequence reconstruction relies on step-by-step textual reasoning, we conduct an ablation study removing the intermediate reasoning traces, forcing the model to generate the final image ordering directly in a single-step reply. As demonstrated in Table~\ref{tab:exp_cot}, removing the CoT leads to a performance drop, demonstrating that the intermediate textual reasoning is crucial to guide and refine the predicted latent sequence $\hat{h}_{\mathrm{final}}$ toward the correct image ordering. \looseness=-1

 \begin{table*}[h]
    \centering
    \caption{\textbf{CoT Ablation Study Results.} We compare the performance of ChronoVision with and without CoT across the In-Domain categories of \benchname{}. Forcing the model to generate the final sequence directly in a single step leads to a consistent performance drop, demonstrating that step-by-step textual reasoning is crucial for guiding the latent sequence generation.}
    \label{tab:exp_cot}
    \setlength{\tabcolsep}{6pt}
    \renewcommand\arraystretch{1.15}
    \resizebox{\linewidth}{!}{%
      \begin{tabular}{lcccccc}   
        \toprule
        \textbf{Variant} & \textbf{Fluid Intel.} & \textbf{Crystallized Intel.} & \textbf{Visuospatial Cog.} & \textbf{Mental Simu.} & \textbf{Transformation} & \textbf{Avg.} \\
        \midrule
        ChronoVision without CoT & 66.2 & 71.1 & 68.3 & 60.0 & 69.1 & 67.6 \\
        \rowcolor{line-blue} ChronoVision with CoT  & \textbf{72.3} & \textbf{77.8} & \textbf{75.0} & \textbf{68.0} & \textbf{78.2} & \textbf{74.8} \\
        \bottomrule
      \end{tabular}
    }
\end{table*}

\subsection{Ablation Study on Reasoning Dependence}
\label{appendix:prefix}

To directly test whether the RVH is merely passively following the textual CoT, we perform an ablation study in which we inject an incorrect intermediate reasoning prefix and force the model to continue generation from that step. As shown in~\ref{tab:exp_prefix}, if the RVH only acted as a passive auxiliary signal, then the model should largely inherit the erroneous reasoning and collapse similarly to the SFT \textit{w/o} head variant. Instead, we observe a much smaller degradation for the full ChronoVision model (SFT with ROI + RL) compared to the baseline. This indicates our training framework internalizes visually grounded constraints into the learned policy, making the model more robust to wrong intermediate reasoning prefixes.

\begin{table*}[h]
    \centering
    \caption{\textbf{Reasoning Dependence Ablation Study Results.} We compare the SFT \textit{w/o} head variant and full ChronoVision (SFT with ROI + RL) under standard inference and when forced to generate from an injected incorrect reasoning prefix. The full model exhibits less performance degradation, demonstrating that our framework internalizes visually grounded constraints to resist incorrect intermediate textual logic.}
    \label{tab:exp_prefix}
    \setlength{\tabcolsep}{6pt}
    \renewcommand\arraystretch{1.15}
    \resizebox{\linewidth}{!}{%
      \begin{tabular}{lcccccc}   
        \toprule
        \textbf{Variant} & \textbf{Fluid Intel.} & \textbf{Crystallized Intel.} & \textbf{Visuospatial Cog.} & \textbf{Mental Simu.} & \textbf{Transformation} & \textbf{Avg.} \\
        \midrule
        SFT \textit{w/o} head (forced incorrect prefix) & 50.8 & 57.8 & 50.0 & 40.0 & 50.9 & 50.8 \\
        \rowcolor{line-blue} SFT \textit{w/o} head (standard inference) & 66.2 & 73.3 & 66.7 & 56.0 & 67.3 & 66.8  \\
        \midrule
        ChronoVision (forced incorrect prefix) & 60.0 & 66.7 & 63.3 & 56.0 & 67.3 & 63.2 \\ 
        \rowcolor{line-blue} ChronoVision (standard inference) & 72.3 & 77.8 & 75.0 & 68.0 & 78.2 & 74.8 \\
        \bottomrule
      \end{tabular}
    }
\end{table*}

\subsection{Ablation Study on Intermediate State Perturbation}
\label{appendix:perturbation}

Following the established methodology~\citep{li2026dynamics}, we test whether each generated latent sequence is necessary by selecting successful reasoning trajectories and injecting random noise into the latent sequence at a specific intermediate step. The model then continues decoding from the perturbed state. We measure the MSE between the predicted intermediate latent sequence $\hat{h}_{\mathrm{final}}$ and the final ground-truth latent sequence $h_{\mathrm{final}}$ at different intermediate reasoning steps. The results are averaged over 10 sampled reasoning trajectories. As shown in Table~\ref{tab:exp_noise}, the trajectory remains stable before the intervention. Once the state is perturbed, downstream latent error increases, and the trajectory fails to reconverge to the clean path. This provides evidence that intermediate latent sequences play a causal role in sustaining the reasoning trajectory, rather than merely being correlated with outcome confidence.

\begin{table}[h]
        \centering
        \setlength{\tabcolsep}{3pt} 
        \renewcommand\arraystretch{1.15}
        \caption{\textbf{Intermediate State Perturbation Ablation Study Results.} We report the MSE between the predicted intermediate latent sequence ($\hat{h}_{\mathrm{final}}$) and the ground-truth final latent sequence ($h_{\mathrm{final}}$) across different reasoning steps. Injecting Gaussian noise at Step 30 disrupts an otherwise successful reasoning trajectory. The perturbed trajectory diverges and fails to reconverge to the clean path. This demonstrates that intermediate latent sequences causally drive and sustain accurate reasoning.}
        \label{tab:exp_noise}
        \resizebox{\linewidth}{!}{%
            \begin{tabular}{lccccc}   
              \toprule
              \textbf{Intervention} & \textbf{Step 0} & \textbf{Step 10} & \textbf{Step 20} & \textbf{Step 30} & \textbf{Step 40} \\
              \midrule
              Add Gaussian Noise @ Step 30 & 0.512 & 0.364 & 0.245 & 0.425 & 0.468 \\
              \rowcolor{line-blue} ChronoVision \textit{w/o} Noise & 0.512 & 0.364 & 0.245 & 0.182 & 0.146 \\
              \bottomrule
            \end{tabular}
        }
\end{table}

\subsection{Ablation Study on Latent Sequence Intervention} 
\label{appendix:intervention}

To further test whether the RVH actively guides reasoning, we conduct an ablation study with latent patching experiment~\citep{liang2026latent}. We intervene during intermediate decoding by swapping the predicted latent sequence between correct and incorrect reasoning runs. We measure the change in the probability of generating the correct frame permutation after the intervention ($\Delta p = p_\mathrm{after} - p_\mathrm{before}$). As shown in Table~\ref{tab:exp_patching}, when we patch a distractor's latent sequence into a correct trajectory, the downstream rationale shifts toward that disaractor's reasoning trajectory with wrong image order. Patching a correct latent sequence into an incorrect run actively shifts the reasoning trajectory toward the correct sequence. This provides a strong evidence that the patched intermediate latent sequence carries vital causal information and actively guides the inference of visual transformations.

\begin{table*}[h]
    \centering
    \caption{\textbf{Latent Sequence Intervention Ablation Study Results.} We report the change in probability ($\Delta p = p_\mathrm{after} - p_\mathrm{before}$) of generating the correct image ordering after intervening on the predicted latent sequence. Patching an incorrect distractor's latent sequence into a correct reasoning trajectory disrupts the generation, while patching a correct latent sequence into an incorrect run successfully guides the trajectory back toward the correct image order.}
    \label{tab:exp_patching}
    \setlength{\tabcolsep}{6pt}
    \renewcommand\arraystretch{1.15}
    \resizebox{\linewidth}{!}{%
      \begin{tabular}{lcccccc}   
        \toprule
        \textbf{Intervention Direction (Source $\rightarrow$ Destination)} & \textbf{Fluid Intel.} & \textbf{Crystallized Intel.} & \textbf{Visuospatial Cog.} & \textbf{Mental Simu.} & \textbf{Transformation} & \textbf{Avg.} \\
        \midrule
        Incorrect $\rightarrow$ Correct Patch & -0.42 & -0.38 & -0.45 & -0.40 & -0.52 & -0.44 \\
        \rowcolor{line-blue} Correct $\rightarrow$ Incorrect Patch & +0.25 & +0.22 & +0.30 & +0.26 & +0.35 & +0.28 \\
        \bottomrule
      \end{tabular}
    }
\end{table*}

\subsection{Ablation Study on Linear Probing} 
\label{appendix:linear}

We also measure how predictive the intermediate latent sequence is of the final sequence permutation at different reasoning steps using a linear probe~\citep{alain2016understanding}. Specifically, we freeze ChronoVision and train a lightweight linear classifier to predict the exact image order only using the intermediate latent sequence $\hat{h}_{\mathrm{final}}$ extracted at different steps of the reasoning process. If the model are taking a shortcut, the latent state would immediately identify the correct order from the start. As shown in Table~\ref{tab:exp_probing}, the accuracy increases gradually and progressively rather than appearing fully formed at early steps. This rules out the shortcut learning hypothesis, providing evidence that the target image order logic is truly constructed step-by-step via latent sequences as the textual reasoning unfolds.

\begin{table}[h]
    \centering
    \caption{\textbf{Linear Probing Ablation Study Results.} We report the accuracy of a linear probe trained to predict the correct image order using only the intermediate latent sequence ($\hat{h}_{\mathrm{final}}$) extracted at different reasoning steps. The probing accuracy on the In-Domain set increases as the reasoning chain unfolds, demonstrating that the temporal logic is constructed step-by-step rather than memorized via a shortcut.}
    \label{tab:exp_probing}
    \setlength{\tabcolsep}{8pt}
    \renewcommand\arraystretch{1.15}
    \resizebox{\linewidth}{!}{%
        \begin{tabular}{lccccc}   
          \toprule
           & \textbf{Step 0} & \textbf{Step 10} & \textbf{Step 20} & \textbf{Step 30} & \textbf{Step 40} \\
          \midrule
          \textbf{ID Acc. (\%)} & 18.4 & 31.6 & 49.2 & 65.5 & 74.1 \\
          \bottomrule
        \end{tabular}
    }
\end{table}

\subsection{Comparison with Related Temporal Reasoning Methods}
\label{appendix_subsec:related_method_comparison}

We compare ChronoVision with R1-VL\citep{zhang2025r1vllearningreasonmultimodal}, VL-Cogito\citep{yuan2025vl}, and Latent Sketchpad\citep{zhang2025latent} using the same Qwen3.5-9B backbone and identical SFT data. ChronoVision obtains the highest overall, ID, and OOD accuracy. R1-VL and VL-Cogito supervise the reasoning process without an explicit target for the final visual state. Latent Sketchpad introduces visual latent representations, but its objective is not tied to the final state that determines the correct ordering. The results suggest that direct supervision of the final-state representation provides useful information for temporal ordering.

\begin{table*}[t]
\centering
\small
\setlength{\tabcolsep}{3.5pt}
\resizebox{\textwidth}{!}
{
\begin{tabular}{lcccccccccccccc}
\toprule
\multirow{2}{*}{\textbf{Method}}
& \multirow{2}{*}{\textbf{Overall}}
& \multicolumn{6}{c}{\textbf{In-Domain}}
& \multicolumn{6}{c}{\textbf{Out-of-Domain}} \\
\cmidrule(lr){3-8}
\cmidrule(lr){9-14}
& & \textbf{Avg.} & \textbf{Flu.} & \textbf{Cry.} & \textbf{Vis.} & \textbf{Men.} & \textbf{Trans.}
& \textbf{Avg.} & \textbf{Flu.} & \textbf{Cry.} & \textbf{Vis.} & \textbf{Men.} & \textbf{Trans.} \\
\midrule
SFT + R1-VL
& 70.4
& 72.0 & 70.8 & 75.6 & 71.7 & 64.0 & 74.5
& 68.8 & 64.4 & 72.5 & 70.0 & 60.0 & 71.0 \\

SFT + VL-Cogito
& 70.8
& 72.4 & 70.8 & 75.6 & 73.3 & 64.0 & 74.5
& 69.2 & 64.4 & 72.5 & 70.0 & 60.0 & 72.0 \\

SFT + Latent Sketchpad
& 69.8
& 71.2 & 69.2 & 75.6 & 71.7 & 64.0 & 72.7
& 68.4 & 64.4 & 70.0 & 70.0 & 60.0 & 71.0 \\

\rowcolor{line-blue} \textbf{ChronoVision}
& \textbf{73.2}
& \textbf{74.8} & \textbf{72.3} & \textbf{77.8} & \textbf{75.0} & \textbf{68.0} & \textbf{78.2}
& \textbf{71.6} & \textbf{66.7} & \textbf{75.0} & \textbf{72.5} & \textbf{64.0} & \textbf{74.0} \\
\bottomrule
\end{tabular}
}
\caption{Comparison with related temporal reasoning methods using the same Qwen3.5-9B backbone and identical SFT data. Accuracy is reported in percentage points. Flu., Cry., Vis., Men., and Trans. denote Fluid Intelligence, Crystallized Intelligence, Visuospatial Cognition, Mental Simulation, and Transformation.}
\label{tab:related_method_comparison}
\end{table*}

\section{Additional Evaluation on Real-World Video Reasoning}
\label{sec:real_world_video}
\subsection{Evaluation on Video-Holmes}
\label{appendix_subsec:video_holmes}

We evaluate ChronoVision on Video-Holmes\citep{cheng2025videoholmesmllmthinklike} to test whether the learned temporal reasoning ability transfers beyond the Vbvr-VQA ordering setting. We uniformly sample 32 frames from each video for ChronoVision. Table~\ref{tab:video_holmes} reports the results across seven reasoning categories. ChronoVision obtains an overall score of 45.89, compared with 42.00 for GPT-4o and 45.00 for Gemini-2.5-Pro. The result indicates that the model can transfer to real-world video reasoning tasks.

\begin{table*}[t]
\centering
\small
\setlength{\tabcolsep}{5pt}
\resizebox{0.9\textwidth}{!}{
\begin{tabular}{lccccccccc}
\toprule
\textbf{Model}
& \textbf{Frames}
& \textbf{SR}
& \textbf{IMC}
& \textbf{TCI}
& \textbf{TA}
& \textbf{MHR}
& \textbf{PAR}
& \textbf{CTI}
& \textbf{Overall} \\
\midrule
Qwen2.5-VL-7B
& 32 & 38.40 & 34.80 & 17.60 & 30.00 & 27.10 & 18.60 & 25.20 & 27.80 \\

Video-R1
& 32 & 48.60 & 41.70 & 28.90 & 34.50 & 31.00 & 33.50 & 35.90 & 36.50 \\

GPT-4o
& 32 & 50.00 & 49.60 & 38.80 & 30.00 & 44.00 & 39.20 & 37.00 & 42.00 \\

Gemini-2.5-Pro
& N/A & 46.60 & 49.30 & 46.90 & \textbf{53.00} & 40.10 & 44.30 & 37.40 & 45.00 \\
\midrule

\rowcolor{line-blue} \textbf{ChronoVision}
& 32 & 48.97 & 48.55 & \textbf{47.99} & 52.00 & \textbf{41.87} & \textbf{44.85} & \textbf{38.89} & \textbf{45.89} \\
\bottomrule
\end{tabular}
}
\caption{Results on Video-Holmes. We report performance across the seven reasoning categories and the overall benchmark score. ChronoVision uses 32 uniformly sampled frames for each video.}
\label{tab:video_holmes}
\end{table*}

\subsection{Evaluation on LongVideo-Reason}
\label{appendix_subsec:longvideo_reason}

We also evaluate ChronoVision on LongVideo-Reason\citep{chen2025scalingrllongvideos}, which measures temporal, goal, plot, and spatial reasoning over long videos. As shown in Table~\ref{tab:longvideo_reason}, ChronoVision reaches an overall score of 74.7. It also obtains the highest score in each of the four reported categories.

\begin{table*}[t]
\centering
\small
\setlength{\tabcolsep}{5pt}
\begin{tabular}{lccccc}
\toprule
\textbf{Model}
& \textbf{Temporal}
& \textbf{Goal}
& \textbf{Plot}
& \textbf{Spatial}
& \textbf{Overall} \\
\midrule
LongVILA-7B
& 58.0 & 80.2 & 57.1 & 46.7 & 62.7 \\

Video-R1-7B
& 61.4 & 85.0 & 62.0 & 58.5 & 68.1 \\

Gemini-1.5-Pro
& 65.4 & 81.9 & 67.8 & 53.3 & 69.3 \\

LongVILA-R1-7B
& 68.1 & 85.7 & 70.6 & 53.3 & 72.0 \\

\rowcolor{line-blue} \textbf{ChronoVision}
& \textbf{72.2}
& \textbf{87.0}
& \textbf{71.9}
& \textbf{59.3}
& \textbf{74.7} \\
\bottomrule
\end{tabular}
\caption{Results on LongVideo-Reason. We report temporal, goal, plot, spatial, and overall benchmark scores.}
\label{tab:longvideo_reason}
\end{table*}

\section{Visualization Example}

\begin{figure*}[h]
    \centering
    \includegraphics[width=\linewidth]{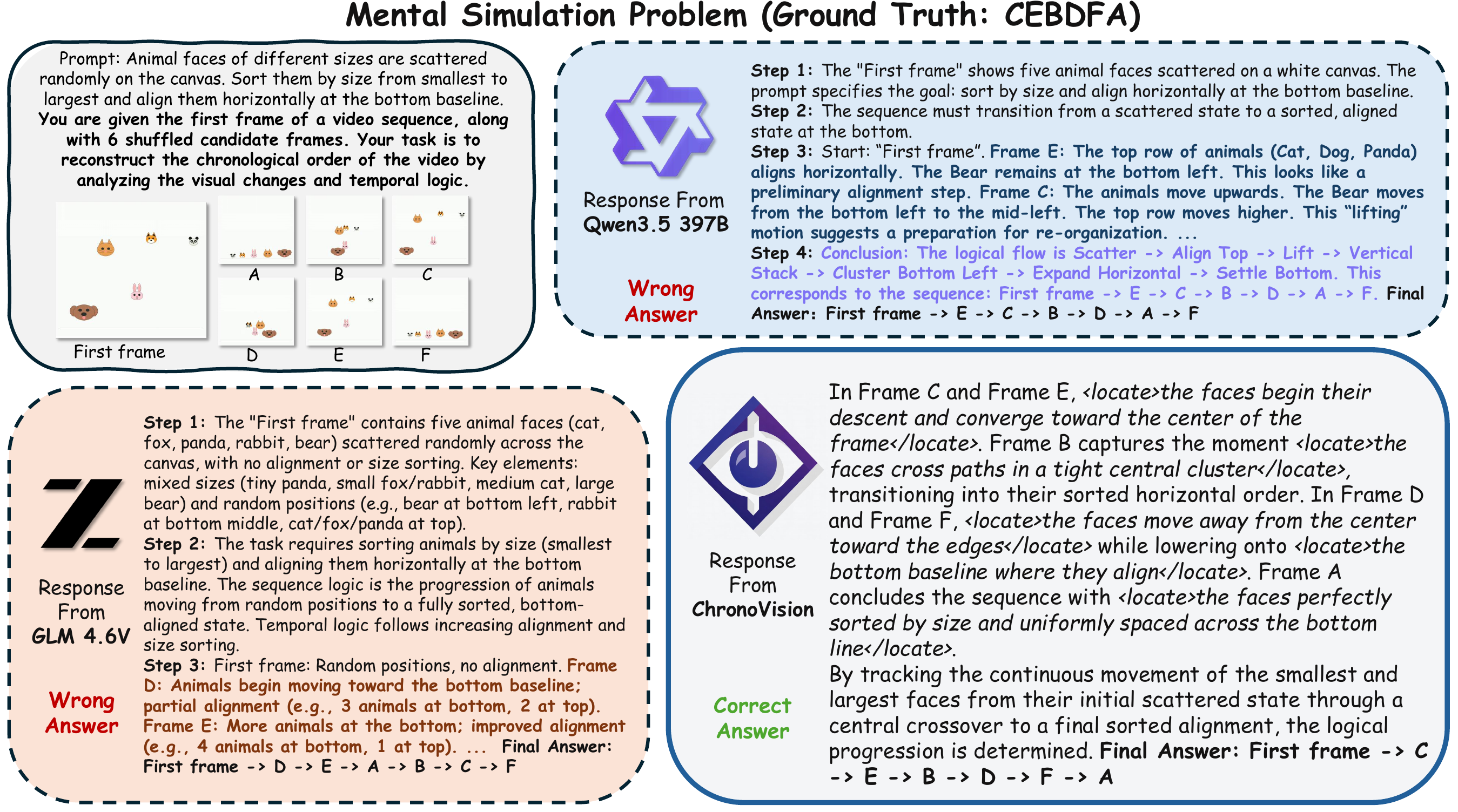}
    \caption{Qualitative comparison on a representative mental simulation problem from Vbvr-VQA. ChronoVision correctly reconstructs the temporal order by grounding reasoning in continuous visual changes.}
    \label{fig:example_vis1}
\end{figure*}

\label{appendix:example_vis}

Figure~\ref{fig:example_vis1} presents a representative mental simulation example from Vbvr-VQA. In this task, the model is required to recover the correct temporal order of six shuffled candidate frames, where a set of animal faces gradually transitions from a scattered initial layout to a horizontally aligned arrangement sorted by size along the bottom baseline. This example highlights that solving the task requires more than static frame matching: the model must track continuous positional changes, intermediate reorganization patterns, and the gradual formation of the target configuration. Compared with strong baselines, ChronoVision better captures this progressive visual evolution and reconstructs the correct sequence by grounding its reasoning in continuous visual changes.

\begin{figure*}[h]
    \centering
    \includegraphics[width=\linewidth]{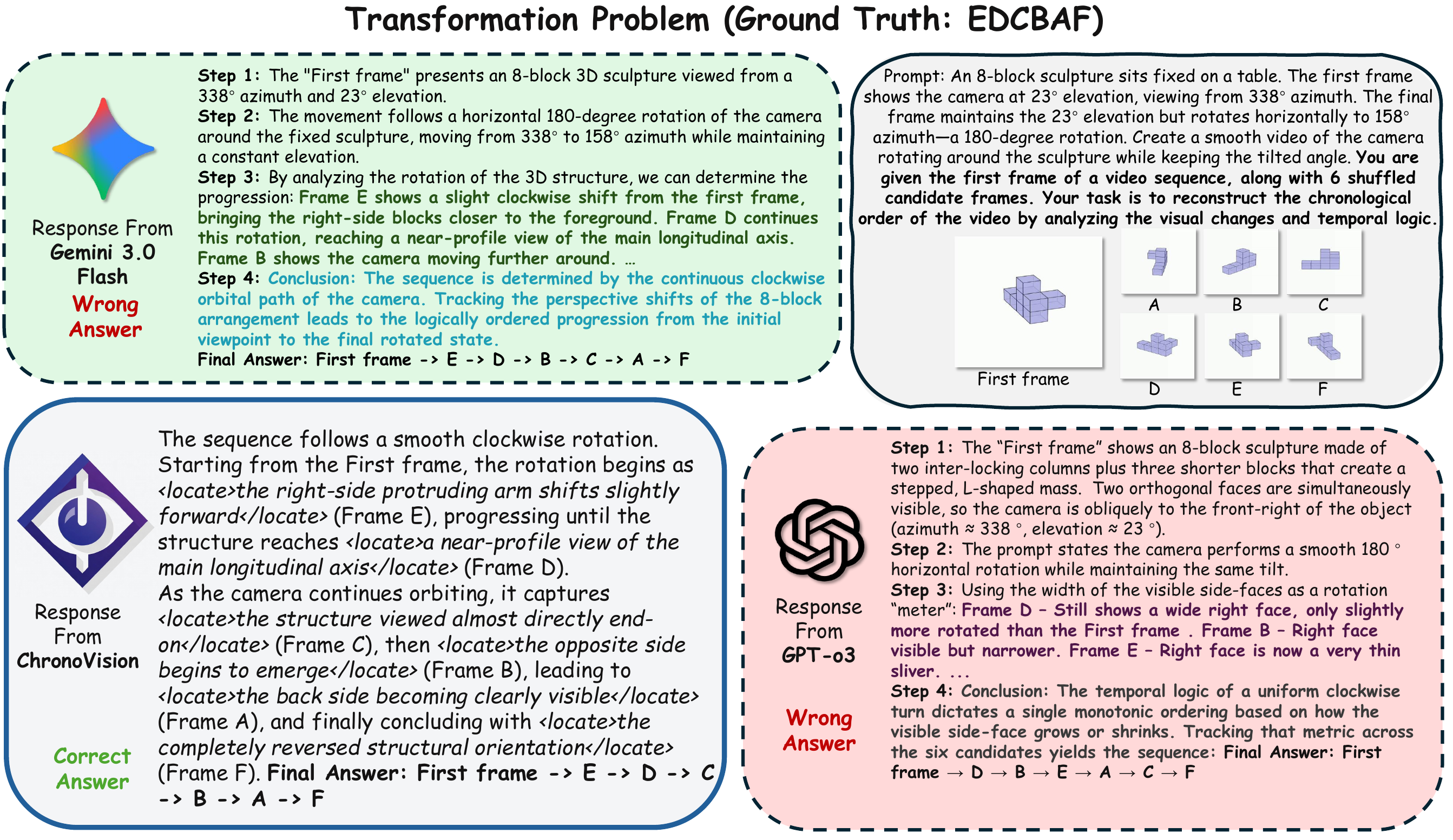}
    \caption{Qualitative comparison on a representative transformation problem from Vbvr-VQA. ChronoVision correctly reconstructs the temporal order by grounding reasoning in continuous structural changes during camera rotation.}
    \label{fig:example_vis2}
\end{figure*}

Figure~\ref{fig:example_vis2} shows a representative transformation example from Vbvr-VQA. Here, the task is to reconstruct the temporal order of a smooth camera orbit around a fixed 8-block sculpture, starting from the initial viewpoint and ending at the reversed orientation after a 180-degree horizontal rotation. Unlike the previous example, this problem mainly tests whether the model can track viewpoint-induced structural changes, including foreshortening, side-face visibility, and the gradual emergence of the opposite side of the object. The qualitative comparison shows that ChronoVision more accurately follows these continuous structural changes and therefore recovers the correct chronological order, whereas baseline models fail to maintain a consistent transformation path.

\section{Related Works}
\label{sec:appendix_related_works}
\subsection{VLMs for Vision Reasoning}
Recent advancements in Vision-Language Models (VLMs) have shifted focus from basic perceptual tasks toward complex deductive reasoning~\citep{ke2025explain,zhang2026cofft,izadi2026visual}. To overcome the text bottleneck, where high-dimensional visual data is compressed into discrete tokens—recent frameworks propose reasoning directly within continuous visual and latent spaces, preserving crucial spatial cues while mitigating hallucinations and over-reliance on superficial language shortcuts~\citep{qin2025chain, zhang2025latent, li2025latent,li2025self}.

Aligning VLMs for protracted reasoning relies heavily on robust Reinforcement Fine-Tuning, notably Group Relative Policy Optimization (GRPO)~\citep{ouyang2022training, shao2024deepseekmath}. Recent works significantly extend this alignment through step-wise rewards, iterative SFT-RL cycles, and progressive curricula to refine complex logic and mitigate reasoning shortcuts~\citep{zhang2025r1vllearningreasonmultimodal, deng2026openvlthinker, xia2025visionary, yuan2025vl}. Furthermore, incorporating fine-grained visual token perception, selective Monte Carlo Tree Search, and self-reflective reasoning traces has proven to drastically enhance data efficiency and spatial grounding during RL optimization~\citep{huang2025spotlight, wang2025sota, wang2025thinknotselectivereasoning, chen2025sft, shen2025fine}.

To bridge perception and logic, visual instruction tuning explicitly aligns visual features with linguistic reasoning~\citep{liu2023visual, huang2025visual}, driving progress across general multimodal benchmarks~\citep{zhang2024mavis} and specialized constrained domains~\citep{zhu2025strive, pham2025rarl}. However, extending these reasoning capabilities to dynamic temporal contexts reveals a persistent gap between static spatial knowledge and true temporal understanding~\citep{feng2025breaking, cores2024lost}. For instance, recent studies on fine-grained action recognition demonstrate that state-of-the-art MLLMs still struggle to decode subtle temporal dynamics, such as pediatric gait behaviors from continuous video sequences~\citep{shen2026decodingchildrensgaitbehavior}. Recent efforts have sought to bridge this gap by enforcing fine-grained motion-language alignment from video clips~\citep{wang2026finemola} and modeling physical dynamics within embodied world simulators~\citep{202606.0173, shen2026egoforge}. To address this temporal bottleneck, our work reformulates video reasoning as an image ordering task to enforce the understanding of visual evolution over time, further grounding the reasoning process by supervising the latent visual representation of the correct final state.

\subsection{Visual Cognition}

Vision-Language Models (VLMs) exhibit profound deficits in core knowledge, intuitive physics, and abstract reasoning compared to human intelligence~\citep{li2025coreknowledgedeficitsmultimodal, schulze2025visual}. This profound discrepancy is fundamentally rooted in the lack of developmental cognitive alignment; unlike human cognition which organically builds core physical concepts during childhood, current VLMs struggle with intuitive physics and require explicit alignment with human cognitive developmental stages~\citep{shen2026position, shen2026evaluating}. Recent studies further highlight a substantial visual cognition gap between humans and multimodal LLMs in abstract visual reasoning~\citep{cao2024visual,li2026toward}. Benchmarks such as MaRs-VQA~\citep{cao2024visual} and VRIQ~\citep{khezresmaeilzadeh2026vriq} reveal that current VLMs are fundamentally bottlenecked by perceptual limitations and multi-image relational cognition rather than language logic alone. Furthermore, comprehensive spatiotemporal assessments including IntPhys 2~\citep{bordes2025intphys}, OmniSpatial~\citep{jia2025omnispatial}, Mmmr~\citep{tie2025mmmr}, and xCrysAlloys~\citep{polat2025stress} demonstrate that VLMs consistently fail at object permanence, spatial dynamics, and physical consistency. To address these deficits, CogSense-Bench introduces a testbed spanning five cognitive dimensions ~\citep{li2026toward}. The Cognitive Supersensing paradigm overcomes text-only reasoning constraints by integrating latent visual imagery prediction, significantly bridging the gap between superficial perception and genuine visual cognition~\citep{li2026toward}.

Matrix reasoning has long served as a core paradigm for studying abstract visual reasoning and its connection to human cognition~\citep{john2003raven, soulieres2009enhanced}. Early work showed that neural models can capture certain compositional visual relations and solve simplified matrix reasoning problems~\citep{fleuret2011comparing, malkinski2025deep, malkinski2023review, xu2023abstract}. Building on this line, benchmark development has played a central role in advancing the field, including RAVEN~\citep{john2003raven}, I-RAVEN~\citep{hu2021stratified}, and CVR~\citep{zerroug2022cvr}, which provide structured testbeds for evaluating relational and analogical reasoning. More recently, RAVEN-IQ~\citep{huang2023language} extends this direction by emphasizing zero-shot visual reasoning settings. By comparison, our work targets this visual cognition gap in temporal reasoning by explicitly training the model to internally simulate the latent visual state of the correct final outcome, rather than inferring it only through textual or symbolic shortcuts.

\subsection{ROI Selection and Cropping}

In visual reasoning tasks, the evidence required to answer a question is often confined to small yet informative regions, while VLMs typically process the entire image at limited resolution, making it difficult to capture dense text, subtle attributes, and fine-grained spatial cues. To address this issue, early studies mainly relied on tool-based ROI selection. Chain-of-Spot \citep{liu2024chain} introduces localized region exploration into multi-step visual reasoning, while ZoomEye \citep{shen2025zoomeye} further performs progressive zooming and region search to acquire key evidence, showing that active local cropping can substantially improve evidence acquisition in complex reasoning. Subsequently, Visual Agents as Fast and Slow Thinkers \citep{sun2024visual} and Refocus \citep{fu2025refocus} incorporate supervised fine-tuning into ROI localization, enhancing local evidence modeling through step-wise decision making and explicit refocusing, respectively. More recent works such as DeepEyes \citep{zheng2025deepeyes} and DeepEyesV2 \citep{hong2025deepeyesv2} further optimize multi-step region exploration with reinforcement learning, enabling models to actively acquire more informative visual evidence during inference. Beyond merely locating regions of interest, it is equally critical to ensure the robustness of these extracted features against pixel-grounding hallucinations~\citep{li2025counterfactual, shen2024practical} and to dynamically route Omni-modal evidence to avoid attention dispersion during multi-step inference~\citep{shen2026cogniroute}.

To avoid the brittleness of explicit coordinate prediction, later work shifted toward attention-driven ROI localization. Methods such as ICoT \citep{gao2025interleaved}, ViCrop \citep{zhang2025mllms}, and FOCUS \citep{zhong2025focus} derive relevant regions directly from internal attention maps rather than autoregressively generating bounding boxes, suggesting that ROI localization is closely tied to the model’s internal ``where-to-look'' signals. However, such signals remain sensitive to layer-wise attention dispersion and to the textual query used for extraction, making stable localization still challenging. Unlike these approaches, our method uses semantically guided ROI attention locating to concentrate internal attention on dynamic key regions, thereby supporting the modeling of temporal visual evolution.

\subsection{Reinforcement Learning}
Reinforcement learning has become an effective paradigm for improving complex reasoning beyond supervised imitation. Recent work has advanced this training pipeline from multiple perspectives, including optimization stability, sampling efficiency, and long-horizon credit assignment. For example, DAPO~\citep{yu2025dapo} stabilizes long-chain reasoning training through decoupled clipping and dynamic sampling, CPPO~\citep{lin2025cppo} improves efficiency by pruning low-contribution completions, and VinePPO~\citep{kazemnejad2024vineppo} improves credit assignment through more accurate return estimation.

On the optimization side, PPO~\citep{schulman2017proximal} remains a standard backbone for post-training, while GRPO~\citep{shao2024deepseekmath} has become a practical choice for reasoning-oriented RL by estimating relative advantages from grouped sampled responses without training a separate value model. Recent analyses further identify biases in GRPO, such as a tendency toward longer responses, motivating refined variants such as Dr.~GRPO~\citep{liu2025understanding}. Unlike prior work that mainly optimizes textual reasoning quality or answer-level outcomes, we adopt GRPO for implicit process grounding in temporal visual reasoning, using a composite reward over sequence correctness, latent visual alignment, and attention focus.

\section{Ethics Statement and Reproducibility}
\label{appendix: G}

\subsection{Artifact License and Data Usage}
The Vbvr-VQA benchmark introduced in this paper is constructed by sampling and processing data from the open-source Very Big Video Reasoning (VBVR) dataset \citep{wang2026bigvideoreasoningsuite}. The research focus and applications of the two are essentially the same, and our use of VBVR is consistent with its intended use. The original VBVR dataset is publicly released under the Apache License 2.0. This license permits free use, modification, and distribution for academic and commercial purposes. Our usage, modifications, and the subsequent release of the Vbvr-VQA task splits strictly comply with the terms of the Apache License 2.0. 

\subsection{Human Validation Subjects and Instructions}
To verify the physical correctness of the ground-truth sequences and the GPT-5 generated reasoning traces, we conducted human validation on the entire testing set (500 samples) and a random subset of the training set (500 samples). 

\textbf{Recruitment and Compensation.} The validation process was performed by graduate-level researchers specializing in computer vision and physics-based machine learning. Since the validation was conducted as an internal academic review procedure to ensure data quality for the benchmark, it did not involve external crowdworkers. Consequently, the reviewers participated voluntarily as part of their academic research duties, and no direct monetary compensation was provided. All participants resided in regions where academic participation of this nature aligns with standard institutional practices. No personally identifiable information was collected. The participants are informed and agree that the results of their validation will be used in this paper.

\textbf{Validation Instructions.} The human reviewers were provided with the following prompt and guidelines via an internal verification interface:

\begin{quote}
\textit{Task Instruction:} You are provided with an initial video frame, a textual task instruction, and six shuffled candidate frames. You are also given the "Ground-Truth Sequence Order" and the "GPT-5 Generated Reasoning Trace." \\
\textit{Verification Rules:} \\
1. Visual Check: Mentally simulate the physical transformation based on the task prompt. Does the provided ground-truth sequence represent a physically plausible and causally valid temporal evolution? (Yes/No) \\
2. Logic Check: Does the provided text rationale accurately explain the temporal progression without hallucinating non-existent objects or violating physical laws? (Yes/No) \\
3. Spatial Check: Do the `<LOCATE>` tags accurately correspond to the key visual regions undergoing dynamic changes? (Yes/No) \\
If any answer is "No," flag the sample for removal or manual correction.
\end{quote}

This strict protocol ensures that the evaluation benchmark is free from synthetic hallucinations and strictly adheres to verifiable physical laws.

\begin{figure*}[p]
\centering
\resizebox{\textwidth}{!}
{
\begin{planbox}{Data Processing Prompt}

\small

You are an expert AI annotator specializing in physical reasoning and
fine-grained spatio-temporal visual tracking. You are provided with the
`First frame`, 6 candidate frames (labeled A through F) depicting a
continuous physical event or transformation, AND the
`Ground Truth Chronological Sequence`
(e.g., First frame -\> C -\> E -\> B -\> D -\> A -\> F).

Your task is to analyze the frames strictly following the provided ground
truth order, identify the key changing objects or dynamic states driving
the temporal progression, and generate dense, region-grounded reasoning
that explains this exact sequence. You must strictly adhere to the
following output structure and rules:

\textbf{Instructions \& Rules:}

\begin{enumerate}[
    label=\arabic*.,
    leftmargin=*,
    labelsep=0.6em,
    itemsep=0.35em,
    topsep=0.35em,
    parsep=0pt,
    align=left
]
    \item \textbf{1. Analyze the Provided Sequence:}
    Briefly explain the overall physical plausibility and causal logic
    governing the transformation from the first frame to the final state,
    based on the given sequence.

    \item \textbf{2. Generate Verb-centric Semantic Cues:}
    You must track the localized dynamic changes across the candidate
    frames. When describing a key visual state or motion that justifies
    a frame's position in the sequence, you MUST enclose the descriptive
    text in \inlinecode{<LOCATE>...</LOCATE>} tags.

    \item \textbf{3. The Action Rule:}
    The text inside the \inlinecode{<LOCATE>} tag MUST be a verb-centric
    phrase describing an action or a changing state
    (e.g., \inlinecode{<LOCATE>the red block being lifted into the air</LOCATE>}
    or \inlinecode{<LOCATE>the faces moving downwards and converging</LOCATE>}).
    Do NOT use static nouns
    (e.g., \inlinecode{<LOCATE>red block</LOCATE>} is invalid).

    \item \textbf{4. Bounding Box Grounding:}
    For every \inlinecode{<LOCATE>} tag you generate, you must output
    the corresponding exact bounding box of that changing/moving object
    in the specified frame, using the format
    \inlinecode{[x\_min, y\_min, width, height]}.
\end{enumerate}

\textbf{Required Output Format:}

\textbf{Step 1: Temporal Logic of the Provided Sequence} \\{}
[Provide the overarching causal logic explaining the given sequence]

\textbf{Step 2: Grounded Visual Cues and Box Coordinates} \\{}

- Following the sequence, Frame [Letter] shows
\inlinecode{<LOCATE>}[verb-centric visual evidence]\inlinecode{</LOCATE>}.
Box: \inlinecode{[x\_min, y\_min, width, height]} \\

- Next, Frame [Letter] captures
\inlinecode{<LOCATE>}[verb-centric visual evidence]\inlinecode{</LOCATE>}.
Box: \inlinecode{[x\_min, y\_min, width, height]} \\

... (Repeat for all key frames in the provided order)

\end{planbox}%
}
\caption{\textbf{Data Processing Prompt.}
The prompt template conditioning GPT-5 on ground truth sequences to
generate verb-centric spatial-temporal reasoning chains and precise
bounding boxes.}
\label{fig:appx_prompt}
\vspace{-0.5cm}
\end{figure*}

\end{document}